\documentclass[preprint,12pt]{elsarticle}

\usepackage{lineno}
\usepackage{amssymb}
\usepackage{amsmath}
\usepackage{comment}
\usepackage{isomath}
\usepackage{bm}
\usepackage{adjustbox}
\usepackage{makecell}

\usepackage{scalerel,stackengine}
\stackMath
\newcommand\reallywidehat[1]{%
\savestack{\tmpbox}{\stretchto{%
  \scaleto{%
    \scalerel*[\widthof{\ensuremath{#1}}]{\kern-.6pt\bigwedge\kern-.6pt}%
    {\rule[-\textheight/2]{1ex}{\textheight}}
  }{\textheight}%
}{0.5ex}}%
\stackon[1pt]{#1}{\tmpbox}%
}

\usepackage{stmaryrd} 
\usepackage[usenames]{color}
\usepackage{epsfig}
\usepackage{graphicx}
\usepackage{psfrag}

\graphicspath{{Figures/}}
\usepackage{float}

\floatstyle{boxed}
\newfloat{algorithm}{htb}{loa}
\floatname{algorithm}{Algorithm}
\usepackage{algorithm,algpseudocode}
\usepackage{caption}
\usepackage{subcaption}
\usepackage{color}
\usepackage[table,xcdraw]{xcolor}
\usepackage{multirow}
\usepackage{natbib}
\modulolinenumbers[5]

\usepackage[left = 2cm, right=2.5cm, top=2cm, bottom =2cm]{geometry}

\begin{document}
\begin{frontmatter}

\title{ Tackling multiphysics problems via finite element–guided \\ physics-informed operator learning }
\author{Yusuke Yamazaki$^{1*}$, Reza Najian Asl$^{2}$, Markus Apel$^{3}$, Mayu Muramatsu$^4$, Shahed Rezaei$^{3*}$}
\address{$^1$Graduate School of Science and Technology, Keio University,\\  Hiyoshi3‑14‑1,Kohoku‑ku, Yokohama 223‑8522, Japan}
\address{$^2$Chair of Structural Analysis, Technical University of Munich,  \\  Arcisstraße 21, 80333 Munich, Germany} 
\address{$^3$ACCESS e.V., Intzestr. 5, 52072 Aachen, Germany}
\address{$^4$Department of Mechanical Engineering, Keio University,\\  Hiyoshi3‑14‑1,Kohoku‑ku, Yokohama 223‑8522, Japan}
\address{$^*$Corresponding authors: yusuke.yamazaki.0615@keio.jp, s.rezaei@access-technology.de}
\begin{abstract}
This work presents a finite element–guided physics-informed operator learning framework for multiphysics problems with coupled partial differential equations (PDEs) on arbitrary domains. The proposed framework learns an operator from the input space to the solution space with a weighted residual formulation based on the finite element method, enabling discretization-independent prediction beyond the training resolution without relying on labeled simulation data. The present framework for multiphysics problems is implemented in Folax, a JAX-based operator-learning platform, and is verified on nonlinear coupled thermo-mechanical problems. Two- and three-dimensional representative volume elements with varying heterogeneous microstructures, and a close-to-reality industrial casting example under varying boundary conditions are investigated as the example problems. We investigate the potential of several neural operators combined with the proposed finite element–guided approach, including Fourier neural operators (FNOs), deep operator networks (DeepONets), and a newly proposed implicit finite operator learning (iFOL) approach based on conditional neural fields. The results demonstrate that FNOs yield highly accurate solution operators on regular domains, where the global features can be efficiently learned in the spectral domain, and iFOL offers efficient parametric operator learning capabilities for complex and irregular geometries. 
Furthermore, studies on training strategies, network decomposition, and training sample quality reveal that a monolithic training strategy using a single network is sufficient for accurate predictions, while training sample quality strongly influences performance. Overall, the present approach highlights the potential of physics-informed operator learning with a finite element–based loss as a unified and scalable approach for coupled multiphysics simulations. 
\end{abstract} 
\begin{keyword} 
Multiphysics problem, Parameterized PDEs, Physics-informed operator learning
\end{keyword}

\end{frontmatter}


\section{Introduction}
Solving partial differential equations (PDEs) along with boundary conditions is fundamental to understanding a wide range of physical phenomena across various engineering disciplines, including structural mechanics, fluid dynamics, and phase transformations in material microstructures. However, the computational cost can become prohibitively high, especially when dealing with large-scale problems involving complex features such as nonlinearity and stiffness. Examples include multiphysics problems, in which multiple governing equations associated with different physical processes are coupled to describe complex phenomena.Thermo-mechanical coupling in the solidification process \cite{Koric2006, zappulla2020multiphysics}, thermo-mechanical phase-field coupling for thermal fracture \cite{RUAN2023105169}, chemo-mechanical coupling \cite{loret2002chemo}, hydro-mechanical coupling for hydraulic fracture \cite{zhou2013new}, and electro-chemo-mechanical coupling in Lithium-ion batteries \cite{hofmann2020electro} are part of the multiphysics examples. 

In recent years, deep learning (DL) has emerged as a promising tool, leveraging the expressive power of neural networks (NNs) to efficiently approximate solutions of PDEs and reduce computational cost. Popular neural network architectures, such as convolutional neural networks (CNNs), recurrent neural networks (RNNs), and graph neural networks (GNNs) have been applied in computational mechanics. Geneva et al.  \cite{geneva2020modeling} proposed an auto-regressive convolutional encoder-decoder architecture for nonlinear spatiotemporal dynamics.
Gao et al. \cite{gao2021phygeonet} introduced a coordinate mapping to CNN to handle irregular domains while preserving their advantages of local feature extraction and parameter efficiency. 3D CNN \cite{pokharel2021physics} and convolutional residual network blocks \cite{liu2024multi} have also been utilized, whose architecture and input/output are tailored to their target PDEs.
Hu et al. \cite{hu2022accelerating} utilized an RNN and dimensionality reduction methods to learn microstructure evolutions described by phase-field models in a latent space. Pfaff et al. \cite{pfaff2020learning} proposed MeshGraphNets, which utilized graph neural networks to learn solutions of mesh-based simulations. Many other studies have presented CNN-based, RNN-based, and GNN-based models for various PDEs, see \cite{qu2022learning,maulik2021reduced,pichi2024graph,franco2023deep} for example.  Furthermore, non-intrusive reduced order modeling for spatiotemporal PDEs, which exploits a latent space constructed by trained autoencoders and latent ordinary differential equations identified through training, has also been explored and shown potential as a fast surrogate while maintaining prediction accuracy in long rollout \cite{fries2022lasdi,bonneville2024gplasdi, longhi2026latent}. 

Physics-informed neural networks (PINNs) are also one of the popular models to approximate PDE solutions with NNs \cite{Raissi.2019}, which is enabled by automatic differentiation in modern deep learning frameworks such as TensorFlow, PyTorch, and JAX. 
The potential of PINNs has been explored through modifying the architecture and training algorithm tailored to specific problems \cite{Mattey.2022,Zobeiry.2021,zhao2024comprehensive}, and PINNs have demonstrated high efficiency and accuracy in solving inverse, ill-posed, and high-dimensional problems that are computationally expensive or unstable for traditional numerical methods \cite{Karniadakis.2021,jagtap2022physics,hu2024tackling,HU2024112495}.
PINNs have also been extended to multiphysics problems and demonstrated promising results in obtaining solutions to coupled and complex systems efficiently.
Chen et al. \cite{CHEN2025118346} presented the effectiveness of a staggered training scheme for coupled systems in phase-field corrosion problems. Harandi et al. \cite{Harandi2023} introduced mixed PINNs to heterogeneous domains in a thermo-mechanical coupling setup. PINNs have been applied to other problem domains such as material degradation \cite{Khadijeh2025} and thermo-chemical curing \cite{AMININIAKI2021113959} with enhanced PINN architecture. The spectral element method is integrated with PINNs to enhance accuracy and efficiency in solving multiphysics problems \cite{SHUKLA2025117498}.

Recently, operator learning, which learns mappings between functions defined on infinite-dimensional Banach spaces, has gained popularity due to its superior capability of predicting PDE solutions across a wide range of problems and resolutions.
Two popular models are Fourier neural operators (FNOs) \cite{li2020fourier,li2021physics,rashid2022learning} and deep operator networks (DeepONets) \cite{Lu2021,Goswami.2022,He.2023b,yin2022simulating}. 
Several other operator learning models have been proposed, such as Laplace neural operator \cite{cao2024laplace}, deep green networks \cite{gin2021deepgreen}, Wavelet neural operator \cite{tripura2023wavelet}, and convolutional neural operator \cite{raonic2023convolutional}. 
The potential of operator learning approaches has been demonstrated in various problem domains, including fluid flows \cite{wen2022u,zheng2025cf}, solid mechanics \cite{li2023fourier,harandi2025spifol,rashid2023revealing}, seismology \cite{haghighat2024deeponet}, phase-field modeling \cite{li2023phase,ciesielski2025deep}, and homogenization \cite{nguyen2025universal}, with the advancement of their architecture and training strategies.
Furthermore, implicit neural representations (INRs), which have initially gained popularity in computer vision have also been extended to continuous learning of PDE solutions due to their superior ability to continuously represent fields. 
Several neural field architectures have been proposed for continuous solutions of spatiotemporal PDEs both in time and space, such as dynamics-aware INR (DINo)\cite{yin2022continuous}, vectorized conditional neural fields (VCNeFs) \cite{hagnberger2024}.
Serrano et al. \cite{serrano2023operator} proposed a coordinate-based model for operator learning (CORAL) for various problem domains such as dynamics modeling and geometry-aware inference, demonstrating its versatility. 
Conditional neural fields can also be combined with latent diffusion models to achieve generative modeling of turbulence flow \cite{du2024conditional}.
Our previous work proposed implicit Finite Operator Learning (iFOL) for continuous solutions of nonlinear PDEs \cite{asl2026physics} on arbitrary geometries. iFOL utilizes a physics-informed encoder–decoder architecture to learn the mapping between continuous parameter spaces and corresponding solution fields. The decoder reconstructs parametric solutions through an implicit neural field conditioned on a latent feature representation. Instance-specific latent codes are obtained via a PDE-constrained encoding procedure formulated as a second-order meta-learning problem. Training relies solely on a physics-informed objective derived from the method of weighted residuals based on the finite element discretization, eliminating the need for labeled data.
Its performance is demonstrated on hyperelasticity, nonlinear thermal diffusion, and phase separation by a phase-field model on both regular and irregular domains. Moreover, iFOL is also combined with the nonlinear FEM to form a hybrid solver, termed Neural initialized Newton, to enhance the convergence of nonlinear FEM \cite{taghikhani2025neural}.

Operator learning models have also been applied to multiphysics problems, with advancements in their architecture. Examples include DeepM\&Mnet on a two-dimensional electroconvection problem for unseen electric potentials \cite{CAI2021110296}, Codomain Attention Neural Operator (CoDA-NO) which demonstrates superior performance on fluid-structure and fluid-thermal interactions \cite{rahman2024}, physics-informed parallel neural operator (PIPNO) \cite{yuan2025}, and coupled multiphysics operator learning (COMPOL) with FNO backbone \cite{li2025multi}.  Systematic evaluations of the performance of neural operators on multiphysics problems have also been conducted to explore their potential in addressing such complex problems \cite{kobayashi2025, yang2025}. 
Their architectures have also been modified to enable accurate predictions of complex physical phenomena modeled by phase-field models, such as fracture \cite{GOSWAMI2022114587} and liquid-metal dealloying \cite{Bonneville2025}.

These studies have demonstrated the applicability of machine learning approaches to multiphysics problem domains. 
However, so far, it has still been limited to relatively simple two-dimensional geometries, and a comprehensive understanding of applying physics-informed loss functions, such as discretized weak form losses, to multiphysics coupled systems in the realm of operator learning has not been well explored. Therefore, it is essential to gain a deeper understanding of the prediction capabilities of neural operators on different problem setups, ranging from a two-dimensional regular domain to an application-oriented three-dimensional irregular domain.
\label{sec:Introduction}
\begin{figure}[t]
    \centering
    \includegraphics[width=0.9\linewidth]{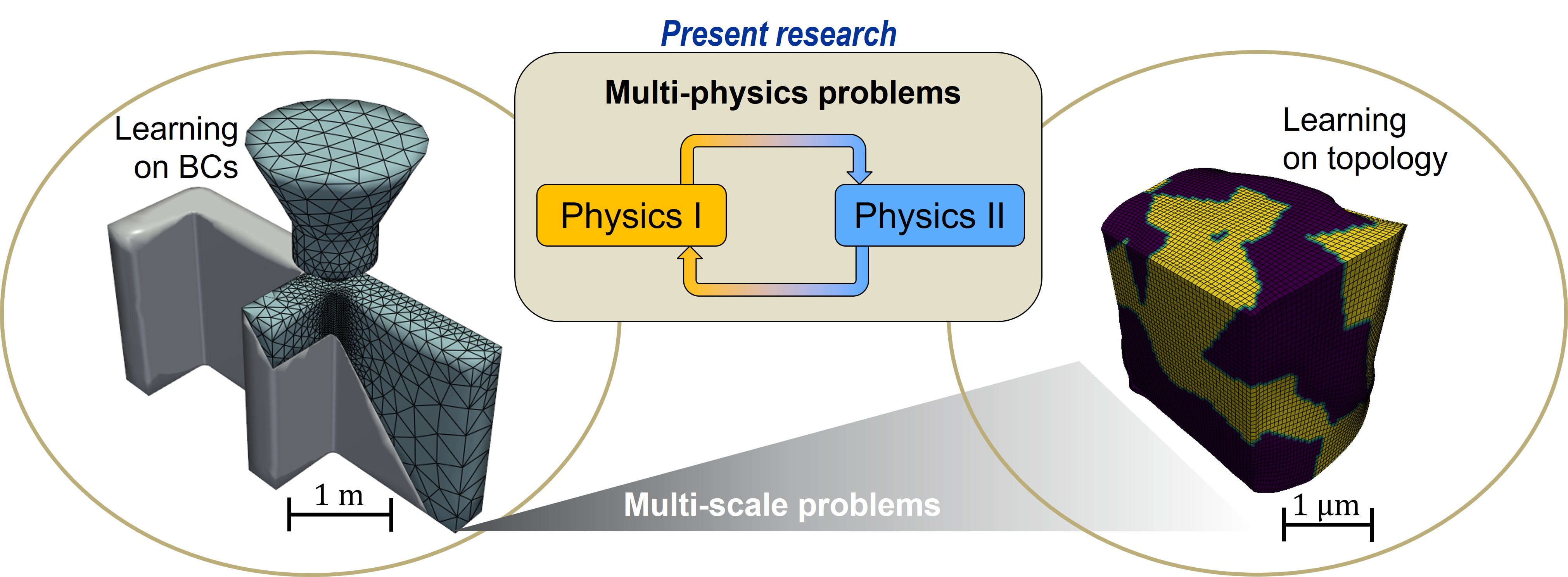}
    \caption{Overall idea of the present research: addressing multiphysics problems across scales with a novel and scalable physics-informed operator learning framework, where finite element residuals are directly used for parametric training of deep learning models. The final goal would be to enable strong coupling of multiphysics problems between macro and microscale as illustrated on the bottom. As a first step, the present work demonstrates the performance of the proposed framework on typical multiphysics problems on both scales.}
    \label{fig:intro}
\end{figure}

This study presents a finite element-guided physics-informed operator learning framework for multiphysics problems involving multiple governing equations with nonlinearity to enable surrogate modeling of such problems, including engineering application scenarios on a three-dimensional domain across scales, without labeled data. The concept of the present contribution is depicted in Fig. \ref{fig:intro}. 
The development is built on the JAX-based platform Folax \cite{folax2025github}, which allows for the seamless integration of classical numerical methods into operator learning architectures. 
In this work, we employ multiple types of backbones, such as FNO, DeepONet, and iFOL, to enable a flexible framework regardless of the target geometry by leveraging the advantages of each backbone.
The proposed approach builds on the concept of Finite Operator Learning (FOL) \cite{rezaei2025finite,rezaei2026finite,yamazaki2025finite} to extend classical numerical solvers toward learning parametric, resolution-independent solution operators. The physics-informed loss function is formulated as weighted residuals based on the finite element method with the predicted solution field acting as the test function, thereby enabling the backpropagation of discrete residuals during the training.
The resulting model is not tied to a single boundary value problem or a fixed discretization, but generalizes across varying input parameters, including material property distributions, topologies, and boundary conditions. Importantly, the learning process does not require labeled solution data, resulting in a fully physics-driven parametric solver that operates independently of precomputed simulations and generalizes across resolutions and problem instances.
The advantages of employing the finite element-based loss function are threefold: (i) it enables the use of unstructured meshes, which are essential for accurately representing complex geometries and capturing localized phenomena in multiphysics problems, (ii) it avoids the need for automatic differentiation to compute partial derivatives in the governing equations, which can be computationally expensive and memory-intensive, especially for complex multiphysics problems, and (iii) the loss formulation based on the weak form of the governing equations allows for a more flexible and stable training process, as it can better handle discontinuities and singularities in the solution, which is common in multiphysics problems, and it provides a unified framework to encode coupling terms and interface conditions between multiple physical fields.
To demonstrate the performance of the proposed framework, nonlinear thermo-mechanical coupling with temperature-dependent material properties is considered. It is noted that, however, the proposed framework is highly flexible and can be extended to other multiphysics problems, such as electro-mechanical and chemo-mechanical coupling, by modifying the loss function according to the problem of interest.
We systematically evaluate the performance of neural operators from various aspects.
The example problems considered in this work include not only a two-dimensional square domain problem, but also a three-dimensional representative volume element (RVE) problem and a industrial-oriented three-dimensional casting example to demonstrate the capability of the present framework in addressing application scenario cases involving large-scale complex geometry. 
For the test cases of regular domain problems, practical-oriented distributions from engineering materials are adopted to evaluate the extrapolation performance in a zero-shot super-resolution setting.
We also consider two training schemes analogous to multiphysics FEM, namely the monolithic and staggered training schemes.  
Additionally, separate and single FNO architectures are compared to investigate the possibility of further enhancement in prediction performance. Other aspects, such as the influence of traning samples and inference cost evaluation are also investigated to gain an deeper understanding of the performance. 

This paper consists of five sections. In Section 2, we introduce the derivation of the loss formulation of the coupled partial differential equations for thermo-mechanical coupling. Section 3 summarizes the neural operators employed in this work and training algorithms. This is followed by Section 4, which presents and discusses the results for three different problems: the two-dimensional square problem, the three-dimensional representative volume element (RVE) problem, and an application-oriented three-dimensional casting example. The conclusion and outlook are provided in Section 5.

\section{Coupled partial differential equations}
In this study, we focus on thermo-mechanical coupling as an example of a coupled system of PDEs. 
More specifically, we consider a nonlinear steady thermo-mechanical problem that couples heat conduction with elastic deformation through thermal strains. The nonlinear behavior mainly originates from the temperature-dependent material properties.
In addition, spatial heterogeneity is incorporated to represent the microstructural features of engineering materials.
Heat transfer is modeled by steady-state conduction, where the thermal conductivity depends on both spatial position and temperature \(k(\boldsymbol{X},T)\):  
\begin{align}
\nabla \cdot  \bm{q}(\boldsymbol{X},T)  &= 0 \quad &&\text{in } \Omega, \\
T &= \bar{T} \quad &&\text{on } \bar{\Gamma}_D, \\
\bm{q} \cdot \bm{n} &= \bar{q} \quad &&\text{on } \bar{\Gamma}_N, \\
\bm{q}(\boldsymbol{X},T) &= - k(\boldsymbol{X},T) \nabla T. &&
\end{align}  
Here, \(T\) is the temperature field, \(\bm{q}\) the heat flux vector, \(\bar{T}\) the prescribed temperature on the Dirichlet boundary \(\bar{\Gamma}_D\), and \(\bar{q}\) the prescribed normal heat flux on the Neumann boundary \(\bar{\Gamma}_N\).  
The mechanical equation with temperature-dependent elastic properties is given by:  
\begin{align}
\nabla \cdot \bm{\sigma}(\boldsymbol{X},T) &= \bm{0} \quad &&\text{in } \Omega, \\
\bm{u} &= \bar{\bm{u}} \quad &&\text{on } \bar{\Gamma}_D, \\
\bm{\sigma} \cdot \bm{n} &= \bar{\bm{t}} \quad &&\text{on } \bar{\Gamma}_N,
\end{align}  
where \(\bm{u}\) denotes the displacement vector, \(\bm{\sigma}\) the Cauchy stress tensor, \(\bar{\bm{u}}\) the prescribed displacement on the Dirichlet boundary, and \(\bar{\bm{t}}\) the prescribed traction on the Neumann boundary.  
Within the thermo-mechanical formulation, the total strain is decomposed into elastic and thermal contributions in an additive manner, i.e. $\bm{\varepsilon} = \bm{\varepsilon}_e + \bm{\varepsilon}_t$, where the thermal strain $\bm{\varepsilon}_t$ reads,
\begin{equation}
\bm{\varepsilon}_t = \alpha \big(T - T_0 \big) \bm{I}.
\end{equation}  
Here, \(\bm{\varepsilon}\) is the total strain tensor, \(\bm{\varepsilon}_e\) the elastic strain, \(\alpha\) the thermal expansion coefficient, and \(T_0\) the reference temperature. In this study, $\alpha$ is set to a constant value for simplicity.
The constitutive relation is described by Hooke's law for heterogeneous linear elasticity:  
\begin{equation}
    \bm{\sigma}(\boldsymbol{X},T) = \mathbb{C}(\boldsymbol{X}) \big( \bm{\varepsilon}(\boldsymbol{X}) - \bm{\varepsilon}_t(\boldsymbol{X},T) \big).
\end{equation}
For isotropic materials, the fourth-order elasticity tensor $\mathbb{C}(\boldsymbol{X})$ is defined in terms of the Young's modulus $E(\mathbf{X})$ and Poisson's ratio $\nu(\mathbf{X})$ as  
\begin{equation}
    \mathbb{C}_{ijkl}(\boldsymbol{X}) 
    = \frac{E(\boldsymbol{X})}{1+\nu(\boldsymbol{X})} 
    \left( \delta_{ik}\delta_{jl} + \delta_{il}\delta_{jk} \right)
    + \frac{E(\boldsymbol{X}) \, \nu(\boldsymbol{X})}{\big(1+\nu(\boldsymbol{X})\big)\big(1-2\nu(\boldsymbol{X})\big)} \, \delta_{ij}\delta_{kl},
\end{equation}
where $\delta_{ij}$ denotes the Kronecker delta. This formulation naturally incorporates material heterogeneity through spatially varying elastic constants over the domain. 

\begin{figure}[t]
    \centering
    \includegraphics[width=0.5\linewidth]{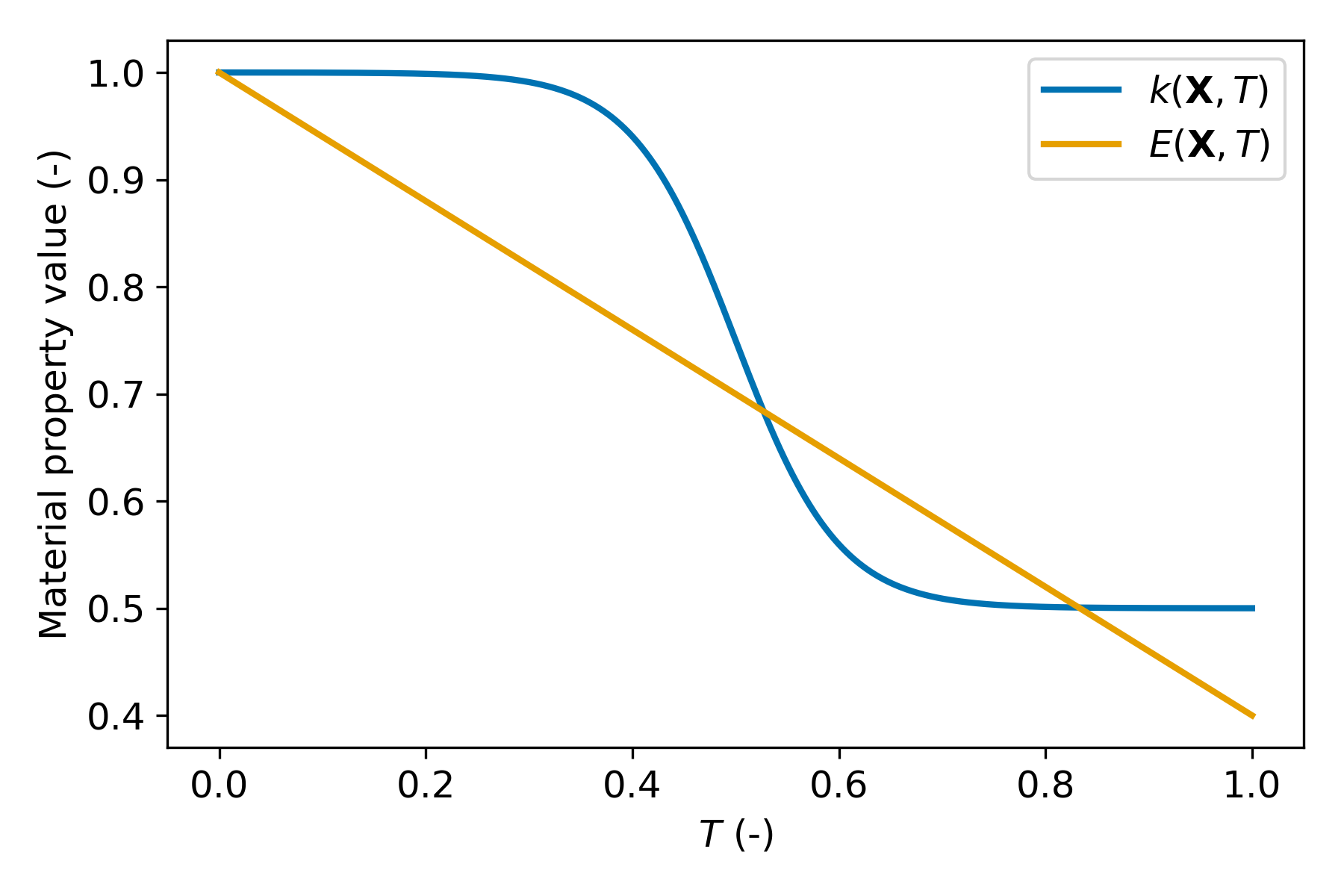}
    \caption{Temperature-dependent material properties utilized in this study.}
    \label{fig:material_property}
\end{figure}

In this study, the following temperature-dependent thermal conductivity and Young's modulus is introduced to account for the nonlinear thermo-mechanical coupling:
\begin{equation}
    k(T,\boldsymbol{X}) = k_0(\boldsymbol{X})\left(\frac{1}{2} + \frac{1}{2\left(1+e^{20(T-\frac{1}{2})}\right)}\right),
\end{equation}
\begin{equation}
    E(T,\boldsymbol{X}) = E_0(\boldsymbol{X}) (1 - \frac{3T}{5}),
\end{equation}
where $k_0$ and $E_0$ are the given phase contrast values distributed over the domain. This function formula is determined based on the material properties reported in aluminum alloy \cite{cox2014dynamic,bakhtiyarov2001electrical}.
The plot of these functions is shown in Fig. \ref{fig:material_property}.

The thermo-mechanically coupled system is enforced using a composite loss function comprising thermal ($\mathcal{L}_{t}$) and mechanical ($\mathcal{L}_{u}$) terms through a weighted residual formulation based on the finite element method with the predicted solution field as the test function \cite{asl2026physics}. This formulation allows for the backpropagation of the residuals during the training process. The loss is expressed as:
\begin{equation}
\label{eq:loss_total}
\mathcal{L} = \mathcal{L}_{t} + \mathcal{L}_{u} = \sum_{e=1}^{n_{el}} \big(\bm T^e_{\theta}\big)^T \bm r^e_t + \sum_{e=1}^{n_{el}} \big(\bm U^e_{\theta}\big)^T \bm r^e_u.
\end{equation}
Here, the thermal and mechanical losses are written for a discretized finite element mesh,
where $\bm r^e_t$ and $\bm r^e_u$ denote the residual vectors of the thermal and mechanical element level, respectively.  The subscript $\theta$ indicates the neural network parameters. and $n_{el}$ is the total number of elements in the training mesh, with $e$ representing the element index.
The thermal and mechanical residuals for element $e$ is given by
\begin{align}
\bm r^e_t &= 
\sum_{k=1}^{n_{int}} W_k
\bm B_t(\bm\xi_k)^T 
\bm N_t(\bm\xi_k) k(\bm\xi_k, \bm T^e_{\theta}) 
\bm B_t(\bm\xi_k) \bm T^e_{\theta} 
- \sum_{k=1}^{n_{int}} W_k \bm N_t(\bm\xi_k)^T \bm f^e, \\
\bm r^e_u &= 
\sum_{k=1}^{n_{int}} W_k \, \bm B_u(\bm\xi_k)^T \bm D(\bm\xi_k, \bm T^e_{\theta}) \Big( \bm B_u(\bm\xi_k) \bm U^e_{\theta} - \alpha \bm N_t(\bm\xi_k) ( \bm T^e_{\theta} - \bm T^e_0) \bm s^e \Big)
- \sum_{k=1}^{n_{int}} W_k \, \bm N_u(\bm\xi_k)^T \bm f^e.
\end{align}
Here, $k(\bm\xi_k, \bm T)$ denotes the temperature-dependent thermal conductivity, and $\bm N_t$ and $\bm N_u$ represents the shape function matrices for the thermal and mechanical fields, respectively. $n_{int}$ is the number of integration points, with $\bm \xi_k$ being the coordinates of the $k$-th integration point in the reference element. $\bm T^e_0$ is the reference temperature vector at the element level, and $\bm f^e$ is the body force vector. $\bm s^e$ is a coupling vector to appropriately map the thermal strain contribution to the mechanical residual.
Furthermore, we define the thermal gradient matrix 
$\bm B_t = \left[\tfrac{d \bm N_t}{d \bm X}\right]$ 
and the strain–displacement matrix 
$\bm B_u = \left[\tfrac{d \bm N_u}{d \bm X}\right]$. 
The weighting factor is expressed as $W_k = w_k \det(\bm J)$, with $w_k$ being the quadrature weight.
For the mechanical residual, $\bm D$ denotes the elastic stiffness matrix. 
The loss function in Eq.~(\ref{eq:loss_total}) is utilized to train the neural operators. Multiple training schemes, namely staggered and monolithic approaches, along with different backbone network architectures, are investigated. Moreover, multiple coupling strategies between the networks are explored for multiphysics coupling, and these aspects are discussed progressively in the following sections.

\section{Neural operators and training algorithm}
\subsection{Neural operators}
\begin{figure}[t]
    \centering
    \includegraphics[width=1.0\linewidth]{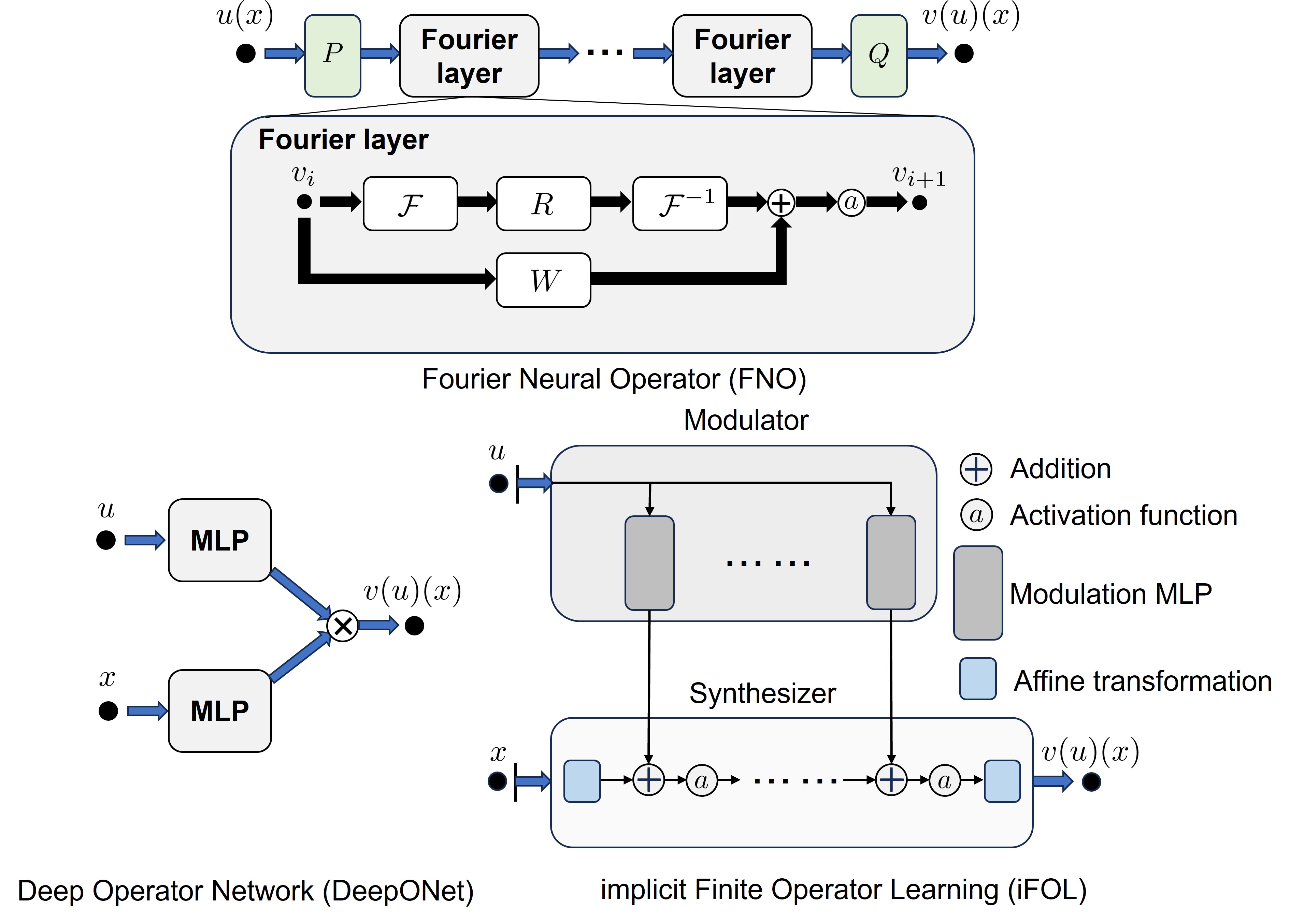}
    \caption{Schematic of the three operator learning backbones employed in this study: Fourier Neural Operator (FNO), Deep Operator Network (DeepONet), and implicit Finite Operator Learning (iFOL).}
    \label{fig:three_OL_schematic}
\end{figure}

We briefly summarize the three neural operator architectures employed in this study: Fourier Neural Operator (FNO) \cite{li2020fourier}, Deep Operator Network (DeepONet) \cite{Lu2021}, and Implicit Finite Operator Learning (iFOL) \cite{asl2026physics}.
The schematic of the three architectures is illustrated in Fig. \ref{fig:three_OL_schematic}. The choice of these architectures is based on the popularity and effectiveness of FNO and DeepONet in learning solution operators of PDEs, as well as the promising performance of iFOL in physics-informed operator learning tasks.
\subsubsection{Fourier neural operator}
The \emph{Fourier Neural Operator} (FNO) is a neural operator that learns mappings between function spaces within the frequency domain using Fourier transforms \cite{li2020fourier}.
Given an input function $u \in \mathcal{U}$ (e.g., material properties or forcing terms), FNO learns an operator $\mathcal{G}: \mathcal{U} \to \mathcal{V}$  such that
\begin{equation}
v(x) = (\mathcal{G}(u))(x),
\label{eq:operator_mapping}
\end{equation}
where  $v \in \mathcal{V}$  denotes the solution field and $x$ is the coordinate.

An FNO layer consists of a spectral convolution combined with a pointwise linear transformation.
For a feature function $v(x)$, the update is written as
\begin{equation}
v_{k+1}(x) = \sigma\left( \mathcal{F}^{-1} \left( R_k \cdot \mathcal{F}(v_k)(\xi) \right)(x) + W_k v_k(x) \right),
\end{equation}
where $\mathcal{F}$ and $\mathcal{F}^{-1}$ denote the Fourier and inverse Fourier transforms, respectively,
$R_k$ is a learnable complex-valued weight tensor acting on a truncated set of Fourier modes $\xi$,
$W_k$ is a local (pointwise) linear operator, and $\sigma(\cdot)$ is a nonlinear activation function. Pratically, the Fourier transform is implemented with the Fast Fourier Transform (FFT) algorithm, which has a computational complexity of $O(N \log N)$ with $N$ being the number of spatial discretization points.
By truncating high-frequency modes, the spectral convolution captures global dependencies efficiently while keeping the computational cost independent of spatial resolution.

\subsubsection{Deep operator network (DeepONet)}
The \emph{Deep Operator Network} (DeepONet) is a neural operator architecture designed to approximate nonlinear operators between function spaces using a two-subnetwork architecture, namely branch-trunk architecture \cite{Lu2021}.
Given an input function $u$ and a query location $x$, DeepONet learns an operator $\mathcal{G}$ satisfying Eq. (\ref{eq:operator_mapping}).
The branch network takes discrete samples of the input function,
\begin{equation}
u = \left( u(x_1), u(x_2), \dots, u(x_m) \right)^{\top}
\end{equation}
and maps them to a latent representation of the input function.
The trunk network takes the query coordinate $x$ as input and produces a location-dependent feature vector.
The choice of the branch network and trunk network architectures is flexible; however, the vanilla DeepONet architecture employs standard multi-layer perceptrons (MLPs) for both networks.
The output of DeepONet is given by an inner product of the branch and trunk features:
\begin{equation}
v(x) = \sum_{i=1}^{p} b_i({u}) \, t_i(x),
\end{equation}
where $\{b_i\}_{i=1}^p$ are the outputs of the branch network and $\{t_i\}_{i=1}^p$ are the outputs of the trunk network.
This structure enables DeepONet to approximate operators by learning a data-driven basis expansion, where the branch network encodes the input function and the trunk network provides coordinate-dependent basis functions.

\subsubsection{implicit Finite Operator Learning (iFOL)}
iFOL builds on conditional neural fields, which employ INRs along with conditioning on input features \cite{asl2026physics} with another network called a modulator. 
While this is in fact analogous to the branch-trunk network architecture in DeepONet, the way of conditioning is enhanced to improve the expressivity of the model.
In this work, we modify the original iFOL architecture to efficiently handle parametric input.
More specifically, we introduce a new modulator network architecture that directly maps the input parameter to the output of the modulator, which acts on the output of each layer of the synthesizer, instead of encoding the input features into latent codes with a PDE-based optimization.
Additionally, skip connections are introduced to the synthesizer to enhance training stability.
Formally, the modulator network at each layer takes the input function or parameter ${u}$ and outputs modulation parameters ${\phi}_i({u})$, and the synthesizer network takes the spatial coordinate $x$ and the modulation parameters to produce the solution field:
\begin{equation}
\begin{aligned}
    v_{\theta} (u) (x) &= {W}_L \left( \sigma_{L-1} \circ \sigma_{L-2} \circ \cdots \circ \sigma_0 ({x}) \right) + {b}_L, \\ 
     \sigma_i({\eta}_i,{\eta}_{i-1},{\phi}_i) & = {\eta}_{i-1}+ a \left( {W}_i \begin{bmatrix} {\eta}_i\\ {\eta}_{i-1} \end{bmatrix}+ {b}_i + {\phi}_i({u}) \right)
\end{aligned}
\end{equation}
\begin{equation}
\begin{aligned}
{\phi}_i({u})
&= {V}_{i,K}\left(\psi_{i,K-1}\circ \psi_{i,K-2}\circ \cdots \circ \psi_{i,0}({u})\right)+{d}_{i,K}\\
\psi_{i,k}({z})
&= a({V}_{i,k}{z}+{d}_{i,k}).
\end{aligned}
\end{equation}
Here, ${\eta}_i$ denotes the output of the $i$-th layer of the synthesizer, ${W}_i$ and ${b}_i$ are the weight matrix and bias vector of the $i$-th layer, respectively.
$a(\cdot)$ is a nonlinear activation function, for which Leaky ReLU is chosen in this work, and $L$ is the total number of layers in the synthesizer.
In the modulator network, ${V}_{i,k}$ and ${d}_{i,k}$ are the weight matrix and bias vector of the $k$-th layer of the modulator corresponding to the $i$-th layer of the synthesizer, respectively, and $K$ is the total number of layers in the modulator.

\subsection{Training algorithm}
Traditional finite element methods for coupled systems employ two main approaches for solving a system of coupled equations: the monolithic scheme and the staggered scheme.
In the monolithic scheme, all governing equations corresponding to different physical fields are assembled into a single global system and solved simultaneously. 
In contrast, the staggered scheme (also referred to as a partitioned or sequential scheme) solves each physical field separately by decoupling the governing equations and iterating between them.
We adapt these two schemes to the training algorithm of neural operators to investigate their effectiveness in learning coupled multiphysics problems in the context of physics-informed operator learning.
In the monolithic approach, the loss of the two governing equations is minimized simultaneously, as shown in Algorithm \ref{alg:training_mnl}, and the staggered approach minimizes each physical loss term from the governing equations alternately, as described in Algorithm \ref{alg:training_stg}, similar to \cite{CHEN2025118346} but in a parametric way.
The switching epoch \(N_n\) in the staggered scheme is a parameter that can be tuned to balance the training with each loss term. 
\begin{algorithm}[t]
\caption{Monolithic training scheme}
\label{alg:training_mnl}
\begin{algorithmic}[1]
\State $n \gets 0$
\While{not converged \textbf{and} $n \le n_{\mathrm{epoch}}$}
    \State $n \gets n + 1$
    \ForAll{mini-batch $\mathcal{M} \subseteq \mathcal{B}$}
        \State $\mathcal{L} = \mathcal{L}_t + \mathcal{L}_m$
        \State $\theta \gets \theta - \lambda \,\frac{1}{|\mathcal{M}|}
        \sum_{i \in \mathcal{M}} \nabla_{\theta}\,
        \mathcal{L}\!\left({v}_{\theta}(x),{u}(x)\right)$
    \EndFor
\EndWhile
\end{algorithmic}
\end{algorithm}

\begin{algorithm}[t]
\caption{Staggered training scheme}
\label{alg:training_stg}
\begin{algorithmic}[1]
\State $n \gets 0$
\While{not converged \textbf{and} $n \le n_{\mathrm{epoch}}$}
    \State $n \gets n + 1$
    \ForAll{mini-batch $\mathcal{M} \subseteq \mathcal{B}$}
        \State \Comment{Switch loss}
        \If{$n \bmod (2N_n) < N_n$}
            \State \text{/*Stage 1 (Thermal)*/}  
            \State $\mathcal{L}_{\text{active}} = \mathcal{L}_t$
            
        \Else
            \State \text{/*Stage 2 (Mechanical)*/}             \State $\mathcal{L}_{\text{active}} = \mathcal{L}_m$
        \EndIf
        \State $\theta \gets \theta - \lambda \,\frac{1}{|\mathcal{M}|}
        \sum_{i \in \mathcal{M}} \nabla_{\theta}\,
        \mathcal{L}_{\text{active}}\!\left({v}_{\theta}(x),{u}(x)\right)$
    \EndFor
\EndWhile
\end{algorithmic}
\end{algorithm}
\section{Results}
\label{sec:Results}
The performance of the multiphysics FOL-based neural operators is presented in this section. To quantitatively evaluate the performance, we carried out studies on several aspects of neural operators on three example problem setups shown in Fig. \ref{fig:domain_bc}.
For simplicity, all physical quantities in this study are normalized to be dimensionless. Nevertheless, the proposed framework can be directly applied to dimensional physical quantities without any modification. We consider the solution by the Nonlinear FEM (NFEM) as the reference solution to evaluate the accuracy of the trained neural operators.

\subsection{Two-dimensional nonlinear thermo-mechanics}
\label{2d_square}
\begin{figure}[t]
    \centering
    \includegraphics[width=1\linewidth]{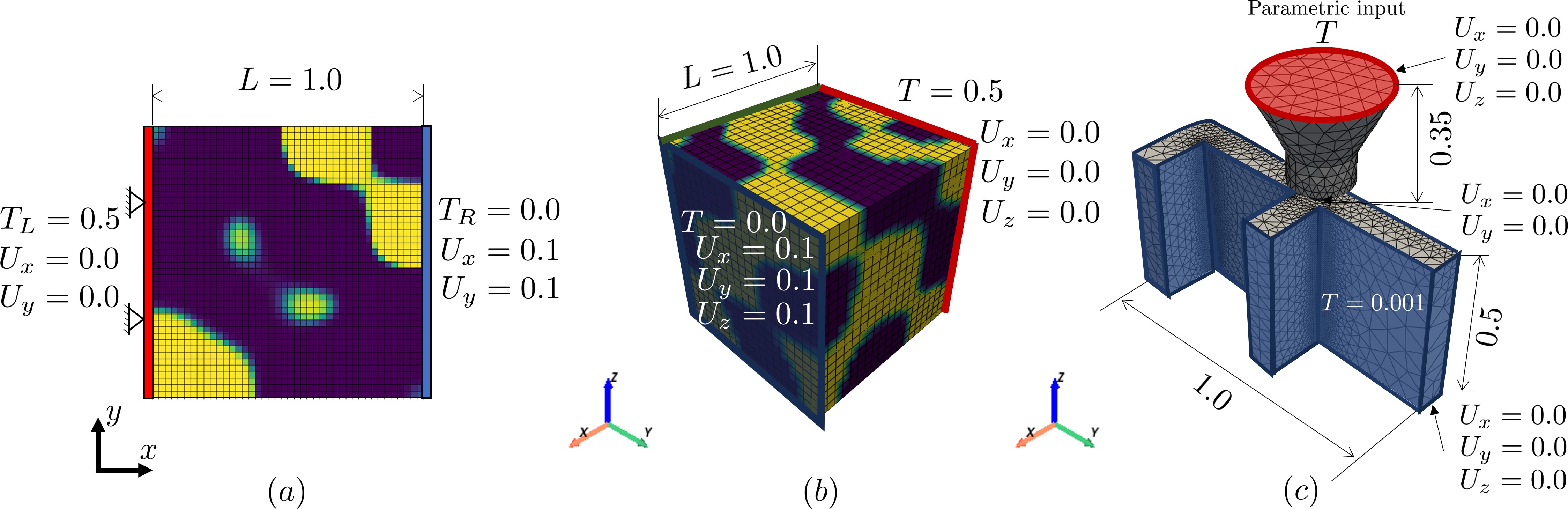}
    \caption{Domain and boundary conditions of the three examples. (a) Two-dimensional squared-domain problem, (b) three-dimensional RVE problem with varying material properties, and (c) casting example problem with varying boundary conditions. }
    \label{fig:domain_bc}
\end{figure}

\begin{figure}[t]
    \centering
    \includegraphics[width=1.0\linewidth]{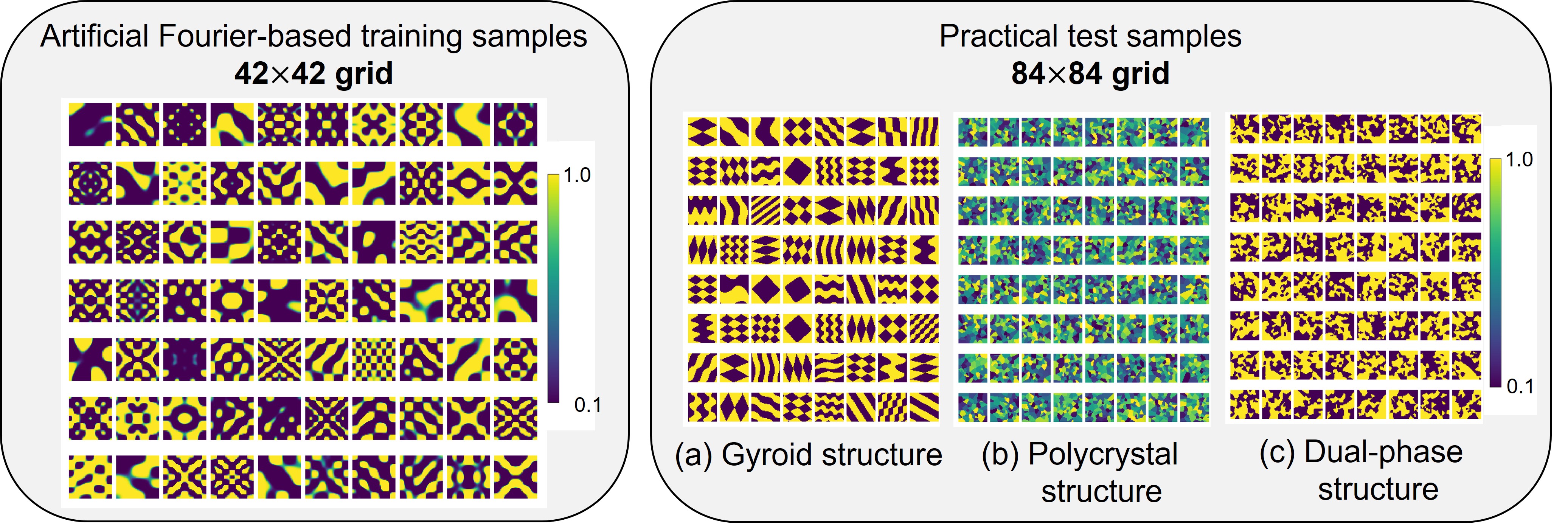}
    \caption{Examples of training (left) and test samples (right) employed for the two-dimensional square-domain problem.}
    \label{fig:train_test_samples}
\end{figure}

We first consider a simple two-dimensional square domain and perform several studies on this problem setup.
The domain and boundary conditions are depicted in Fig.~\ref{fig:domain_bc} (a). 
The training hyperparameters are summarized in Table \ref{tab:hyperparameters_fno}, which is determined based on hyperparameter studies.
For training the neural operator, we generate training samples using the Fourier-based function described in \ref{appendix:sample_generation}. The training samples consist of heterogeneous microstructures with spatially and continuously varying material properties, i.e., thermal conductivity and elastic modulus, as shown on the left of Fig. \ref{fig:train_test_samples}.
For this problem on a regular domain, we employ FNO as a backbone for learning solutions given the microstructure that represents the phase contrasts of material properties, i.e. thermal conductivity and elastic modulus. We refer to FOL with a FNO backbone as FNO-FOL in this paper.
In total, 5000 samples are utilized for training the neural operator model. 


\subsubsection{Studies on prediction accuracy for unseen cases}
\begin{figure}[H]
    \centering
    \begin{minipage}{0.45\linewidth}
        \centering
        \includegraphics[width=\linewidth]{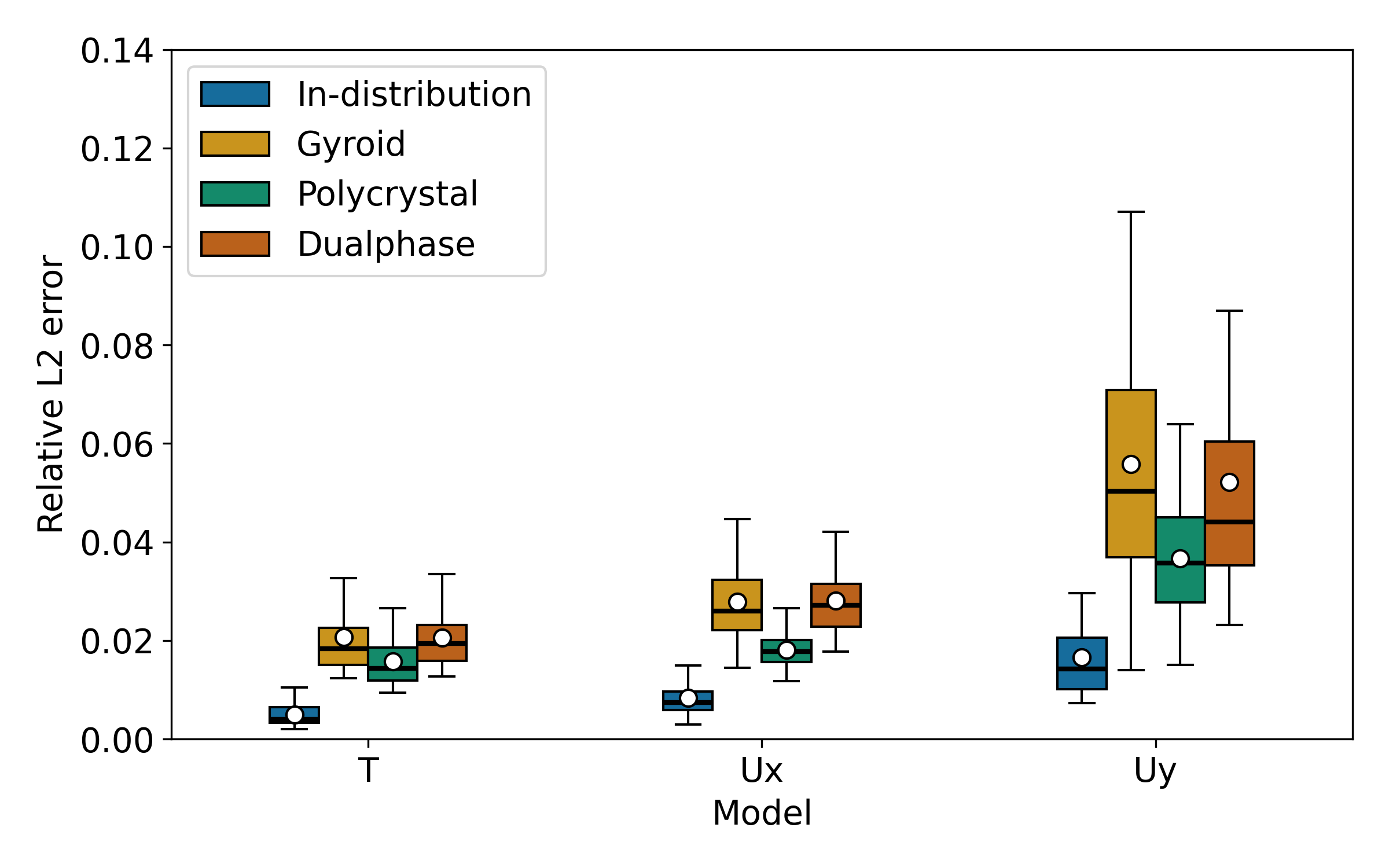}
        \caption{Comparison of the relative L2 errors over 50 samples on four different test cases.}
    \label{fig:comparison_rel_l2_2d}
    \end{minipage}
    \hfill
    \begin{minipage}{0.53\linewidth}
        \centering
        \includegraphics[width=\linewidth]{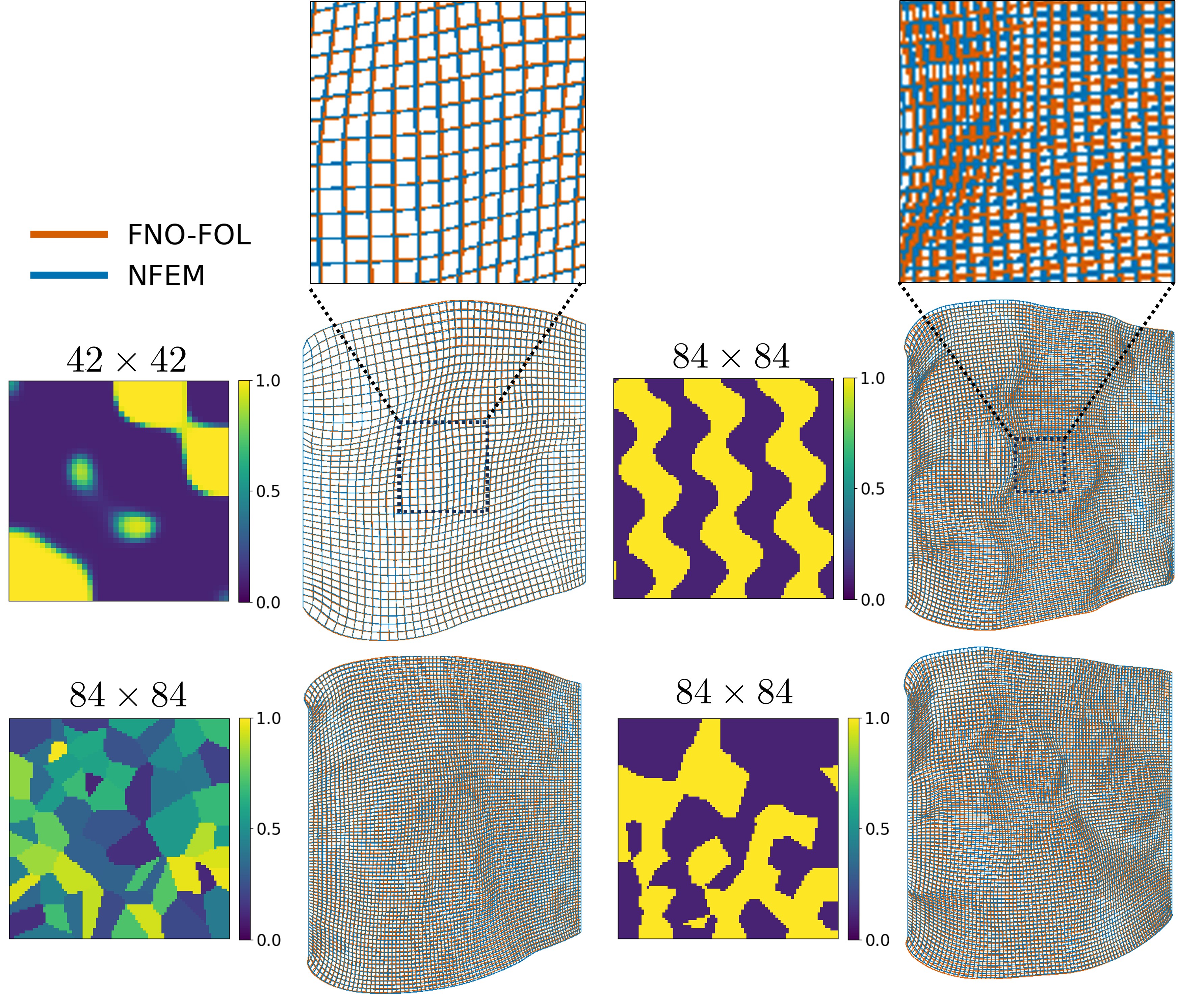}
        \caption{Comparison of the deformed states between prediction and reference solution from NFEM on a representative sample of all test cases.}
    \label{fig:2d_deformed_state}
    \end{minipage}
\end{figure}

\begin{figure}[H]
    \centering
    \includegraphics[width=1.0\linewidth]{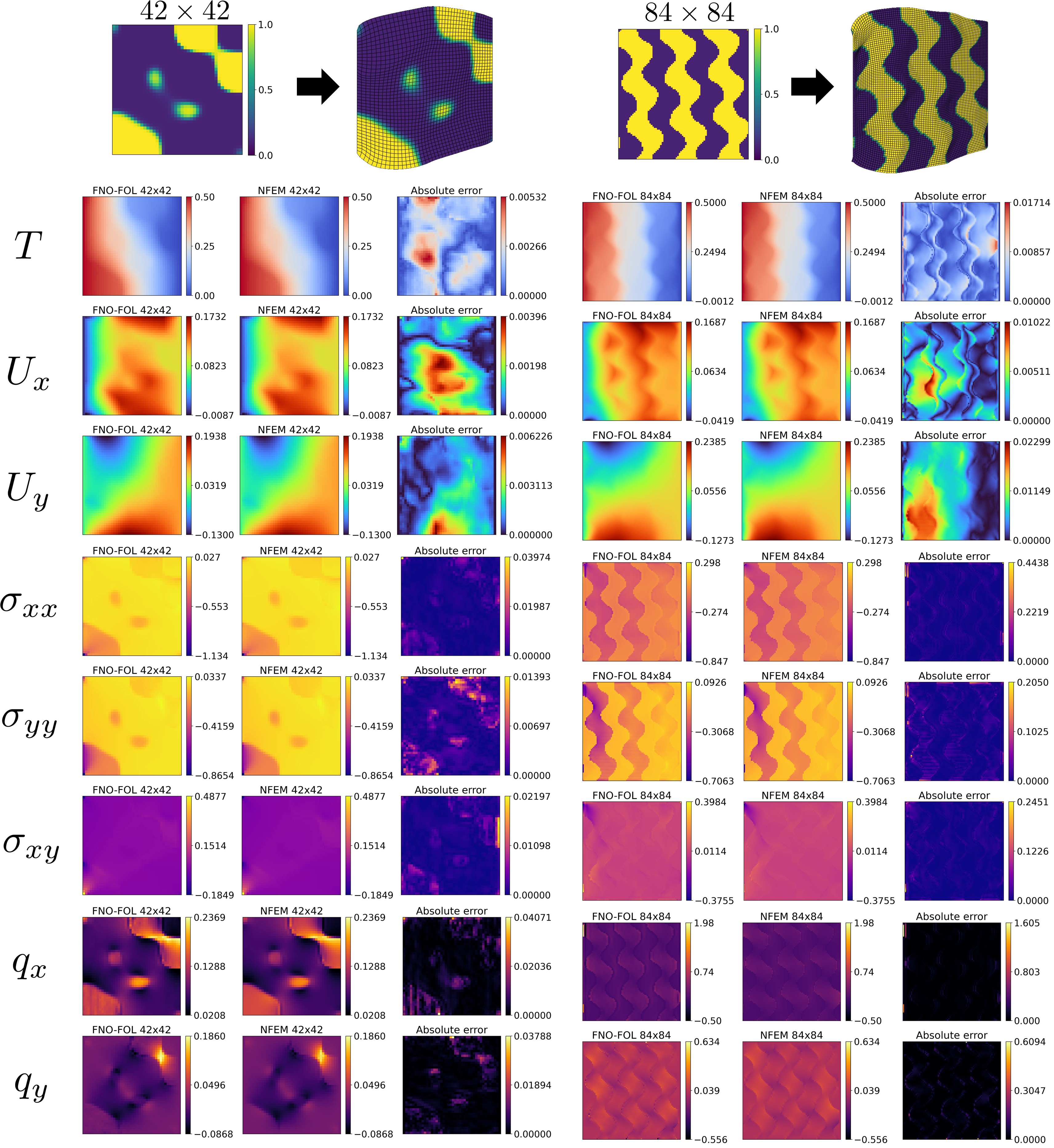}
    \caption{Comparison of the prediction and reference solution from NFEM in the two-dimensional square domain problem for a representative case of the in-distribution and gyroid test samples.}
    \label{fig:results_in_meta}
\end{figure}

\begin{figure}[H]
    \centering
    \includegraphics[width=1.0\linewidth]{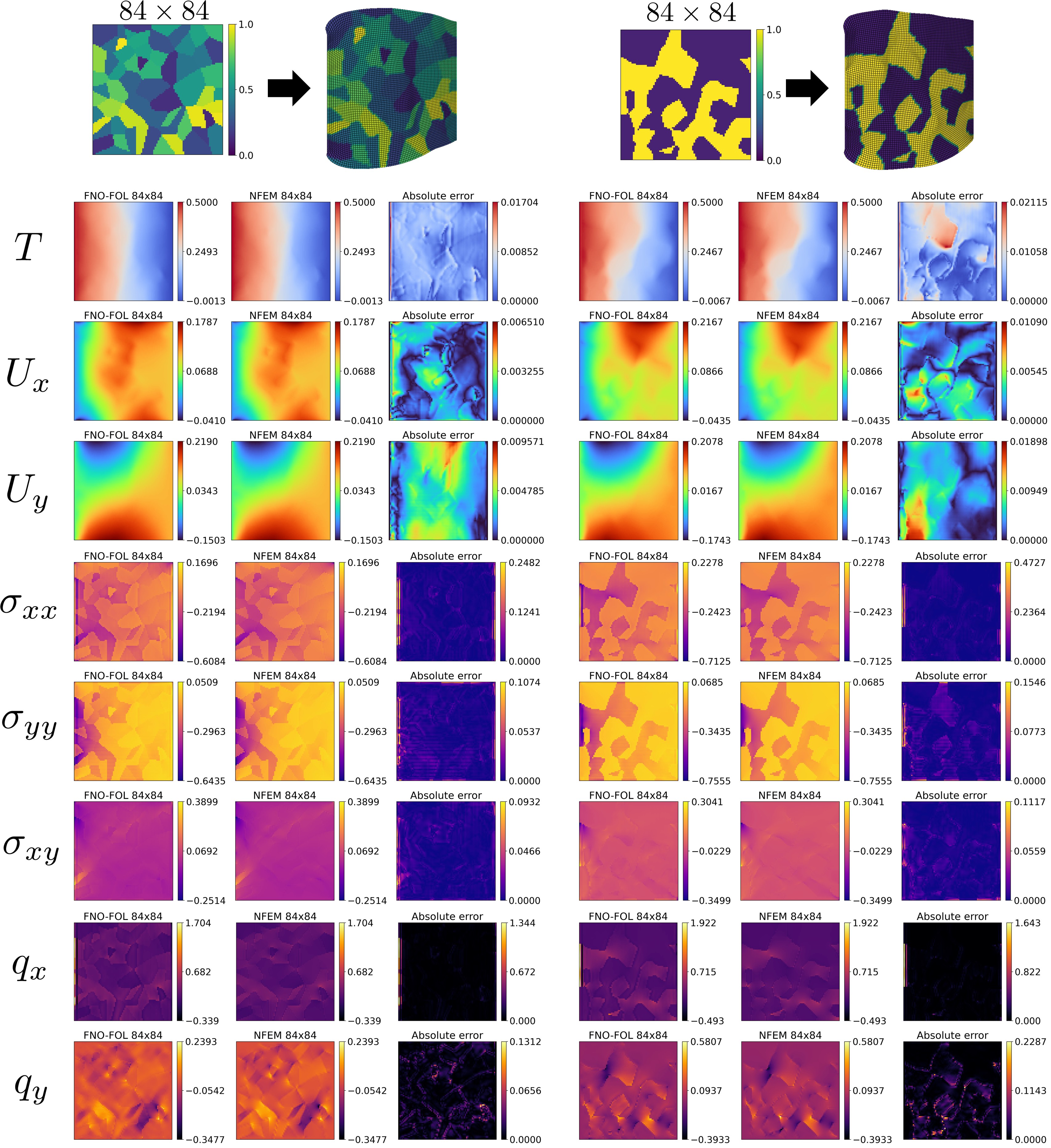}
    \caption{Comparison of the prediction and reference solution from NFEM in the two-dimensional square domain problem for a representative case of the polycrystal and dualphase test samples.}
    \label{fig:results_poly_dualphase}
\end{figure}

In addition to the in-distribution test samples from the Fourier series-based sample generator, we introduce three extreme test samples to evaluate the prediction capability to unseen input distributions.
These test samples include (a) gyroid structures which mimics the microstructure of metamaterials, (2) polycrystalline microstructures generated by Voronoi tessellation, and (3) dual-phase microstructures with sharp interfaces between the two phases. Examples of the test samples are shown in Fig. \ref{fig:train_test_samples}. These samples can be generated using a simple Python script within a few seconds. 
Furthermore, these test cases are generated at a resolution of $84\times 84$, which is finer than the training grid of $42\times 42$ to demonstrate the capability of the trained model to perform zero-shot super-resolution tasks.
In this study, the relative L2 error statistics over 50 samples are computed by taking the relative L2 error between the predicted and reference solution from NFEM for each physical field on each input sample and then averaging the error over 50 samples for each test case. The comparison on the in-distribution test case and the three test cases are shown in Fig. \ref{fig:comparison_rel_l2_2d}, ane examples of the prediction results are shown in Figs. \ref{fig:2d_deformed_state}, \ref{fig:results_in_meta} and \ref{fig:results_poly_dualphase}.
The relative L2 error for all physical fields remains below $3\%$ for the in-distribution test samples and below $10\%$ for the three extreme test cases, demonstrating the accuracy and generalizability of the multiphysics FOL-based neural operator for unseen input distributions. The error for the displacement field $U_y$ is slightly higher than that of other fields, which can be attributed to the smaller magnitude of some displacement values on the domain compared to other physical fields, leading to a larger relative error.
Nevertheless, the deformed configurations compared in Fig. \ref{fig:2d_deformed_state} indicate that there is a good qualitative agreement between the predicted and reference deformed states on every test case. 

Looking at the distribution of the error in Figs. \ref{fig:results_in_meta} and \ref{fig:results_poly_dualphase}, the maximum point-wise error also remains below $10\%$ for all physical fields, indicating that the trained neural operator can accurately capture the spatial distribution of the coupled thermo-mechanical fields even in extreme cases with complex microstructures.
On the other hand, in the three extreme test cases in Figs. \ref{fig:results_in_meta} and \ref{fig:results_poly_dualphase}, it can be observed that the error tends to be higher around the regions with high gradients, such as phase interfaces or regions with rapid changes in material properties. This behavior is particularly pronounced near the Dirichlet boundary, where steep temperature gradients introduce high-frequency components. Due to the spectral truncation inherent in FNO, these high-frequency features are not fully resolved, leading to increased errors in the boundary region, even though the zero-padding technique is employed to mitigate boundary artifacts. 
Addressing this limitation may require incorporating adaptive basis representations or hybrid architectures to better capture fine-scale features.
Furthermore, exploring alternative operator learning architectures that are truly representation-equivalent could enhance the model's performance in these challenging regions.
\noindent
Another point worth mentioning for this phenomenon is this issue would not arise if periodic boundary conditions, as employed in homogenization methods, were considered. However, in this work, we focus on simple Dirichlet boundary conditions to demonstrate the capability of the proposed framework in prediction solution fields under non-periodic boundary conditions.


\subsubsection{Studies on network decomposition for field-specific learning}
\begin{figure}[t]
    \centering
    \includegraphics[width=0.9\linewidth]{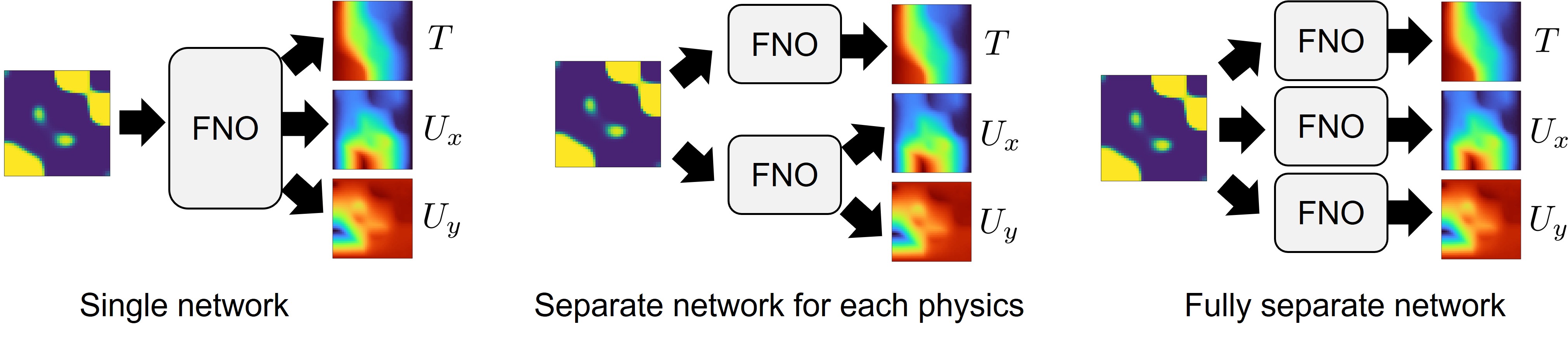}
    \caption{Comparison of the three different network decomposition strategies for FNO-based multiphysics operator learning.}
    \label{fig:network_decomp}
\end{figure}

\begin{figure}[t]
    \centering
    \includegraphics[width=1.0\linewidth]{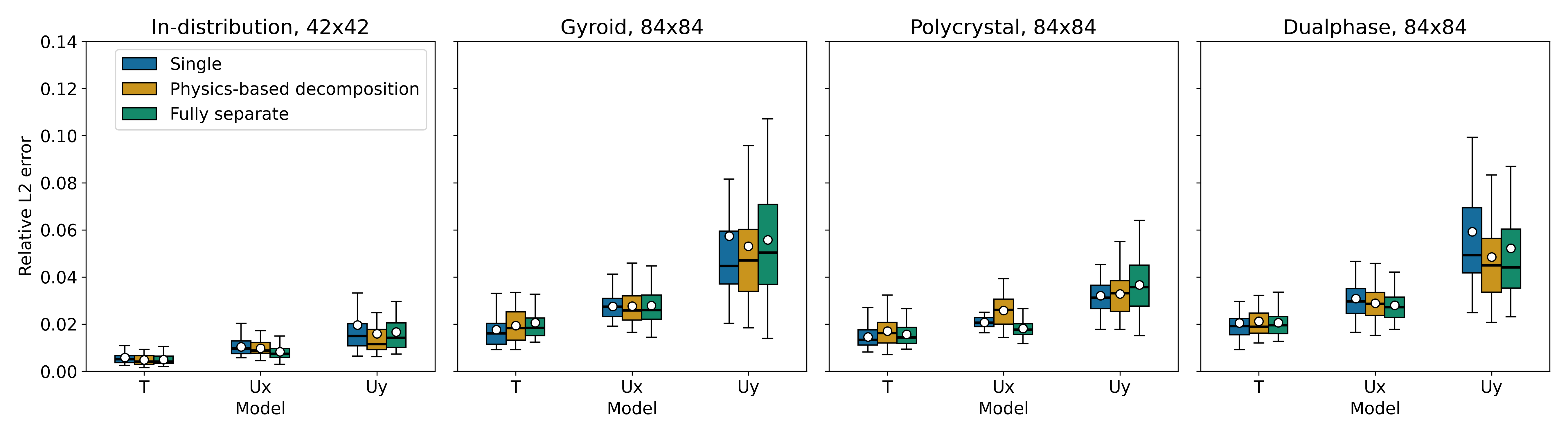}
    \caption{Relative L2 error statistics over 50 samples on four different test cases with three different network decomposition strategies. }
    \label{fig:arch_rel_l2}
\end{figure}

Since each physical field in multiphysics problems may exhibit different characteristics and magnitudes, we investigate different network decomposition strategies to enhance the learning capability of FNO for multiphysics problems.
To that end, three different network decomposition strategies are considered for FNO-based multiphysics operator learning, as illustrated in Fig.~\ref{fig:network_decomp}: (a) a single FNO that learns all physical fields simultaneously, (b) separate FNOs for each physics, and (c) fully separate networks for each physical field.
The training is performed using the same training dataset and hyperparameters as in \ref{appendix:hyperparameters} for all three strategies, except for the number of channels to make the number of trainable parameters comparable, for a fair comparison.
As a result, the relative L2 error statistics over 50 samples on four different test cases with the three different network decomposition strategies in Fig. \ref{fig:arch_rel_l2} demonstrates that while the separate network architecture performs slightly better than the shared architecture for in-distribution cases, the single FNO architecture performs slightly better than the separate FNOs for other cases, especially the polycrystal test case. Overall, the performance difference among the three strategies is not substantial, indicating that FNO can effectively learn multiphysics problems regardless of the network decomposition strategy using the proposed FOL approach
On the other hand,  the training time for the single FNO architecture is approximately 1.34 times faster than that of the separate FNOs and 1.46 times faster than that of the fully separate networks under the comparable number of trainable parameters.
From the perspective of training efficiency, the single FNO architecture could be a preferable choice for multiphysics FOL under the current settings.
To further enhance the learning capability, more advanced architectures such as connection between different FNOs for corresponding fields to enable information exchange among different physical fields could be explored in future work.

\subsubsection{Studies on training schemes: staggered vs. monolithic}
\begin{figure}[t]
    \centering
    \includegraphics[width=1.0\linewidth]{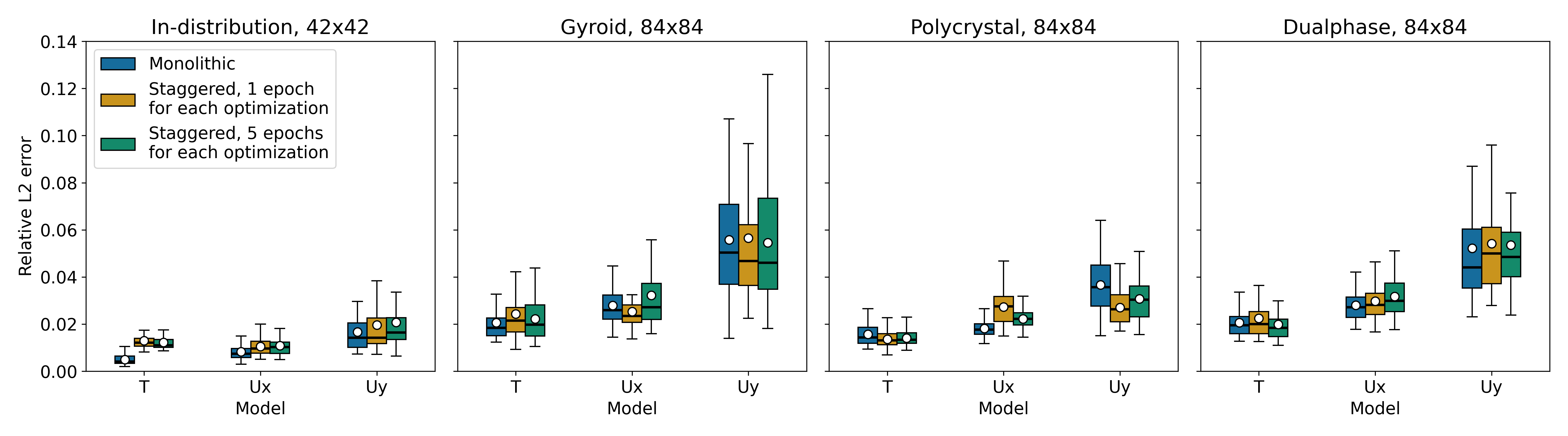}
    \caption{Relative L2 error statistics over 50 samples on four different test cases with the monolithic and staggered training schemes. }
    \label{fig:training_scheme_rel_l2}
\end{figure}

We have introduced two training schemes, the monolithic training scheme in Algorithm \ref{alg:training_mnl} and the staggered training scheme in Algorithm \ref{alg:training_stg} to clarify which approach results in better prediction accuracy for unseen cases, by analogy with the approach in classical numerical methods for coupled systems.
To that end, we trained the same network using the two training schemes and evaluate the prediction accuracy on the same test cases.
As shown in Fig. \ref{fig:training_scheme_rel_l2}, the monolithic training scheme outperforms the staggered training scheme in the in-distribution test case or all the three solution fields. On the other hand, the staggered training scheme, using either 1 or 5 epochs for alternating the loss, performs slightly better in some of the solution fields for the three extreme test cases, although the performance difference is not substantial. 
This could be attributed to the fact that the staggered training scheme allows each physical field to be learned more independently, which may change how the model generalizes to extreme cases in the optimization landscape.
In terms of the training time, the monolithic training scheme are by 2.53 percent faster than the staggered scheme cases, which is attributed to the necessity of the loss alternation in the training loop of the staggered scheme. 
Overall, both training schemes demonstrate comparable performance, indicating that the proposed multiphysics FOL-based neural operator can be effectively trained using either approach. However, taking the training efficiency and algorithmic simplicity into account, the monolithic training scheme could be a preferable choice under the current settings. Therefore, the monolithic training scheme is employed for the rest of the studies in this paper.

\subsubsection{Studies on number and quality of the initial training fields}
\label{sec:training_samples_study}
\begin{figure}[t]
    \centering
    \includegraphics[width=1.0\linewidth]{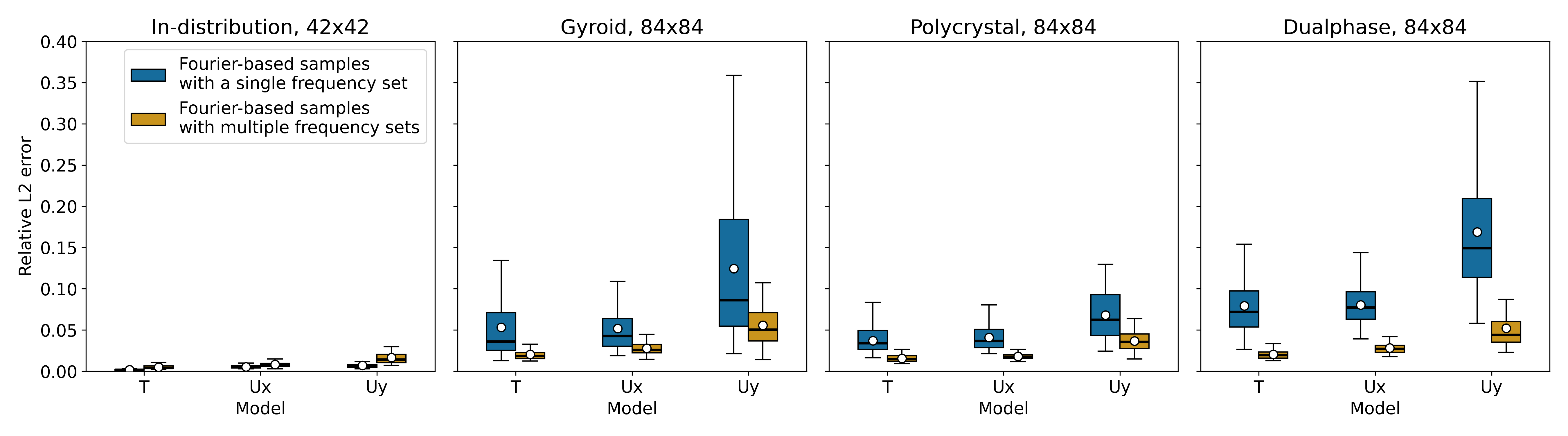}
    \caption{Relative L2 error statistics over 50 samples on four different test cases on two different training samples with different frequency sets in the generation of training samples based on the Fourier series. }
    \label{fig:sample_rel_l2_error}
\end{figure}

\begin{figure}[t]
    \centering
    \includegraphics[width=1.0\linewidth]{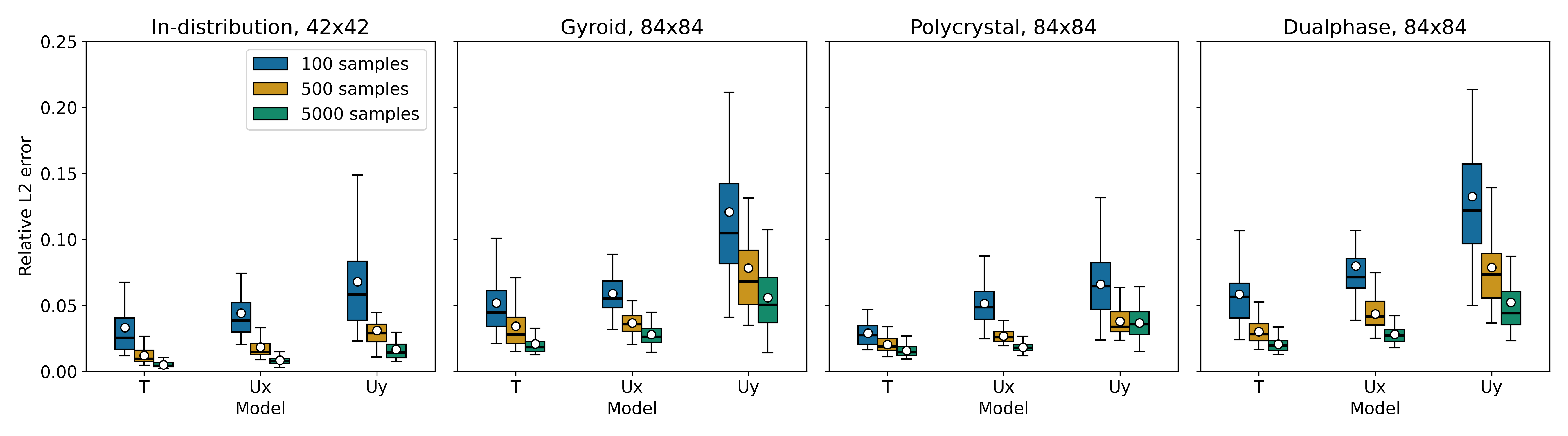}
    \caption{Relative L2 error statistics over 50 samples on four different test cases on two different training samples with different numbers of training samples. }
    \label{fig:number_samples_rel_l2_error}
\end{figure}

We also investigate the influence of the training samples on the prediction accuracy. To that end, we focus on two aspects of the sample generation strategy: the variety of training samples and number of training samples. For the variety of training samples, we generate two different sets of training samples using different frequency components in the Fourier-based sample generator described in \ref{appendix:sample_generation}. The first set uses a single set of frequency components, while the second set incorporates four sets of frequencies with more various frequency components to introduce more complex spatial variations in material properties. For the fair comparison, we use the same number of training samples (5000 samples) for both sets.
As shown in Fig. \ref{fig:sample_rel_l2_error}, the training samples with more various frequency components result in enhanced prediction accuracy for all test cases, reducing the prediction error by more than 50\%. 
This indicates that the variety of
training samples plays a crucial role in enhancing the generalizability of the trained neural operator.
Next, we investigate the influence of the number of training samples on the prediction accuracy. We generate three different training datasets with 100, 500, and 5000 samples using the multiple frequency sets determined based on the training sample study in the Fourier-based sample generator.
As shown in Fig. \ref{fig:number_samples_rel_l2_error}, using more training samples such as 500 and 5000 results in better prediction accuracy for all test cases compared to the 100-sample dataset, demonstrating the importance of a sufficient number of training samples to capture the complex relationships between input material properties and output physical fields.
The results also indicate that increasing the number of training samples leads to a more pronounced improvement in prediction accuracy for extreme test cases compared to in-distribution test cases, suggesting that a larger dataset is particularly beneficial for enhancing generalizability to unseen input distributions within the capacity of the model. 
Overall, these studies highlight the significance of both the variety and quantity of training samples in achieving accurate predictions.

\subsubsection{Inference cost}
\begin{figure}[t]
    \centering
    \includegraphics[width=0.8\linewidth]{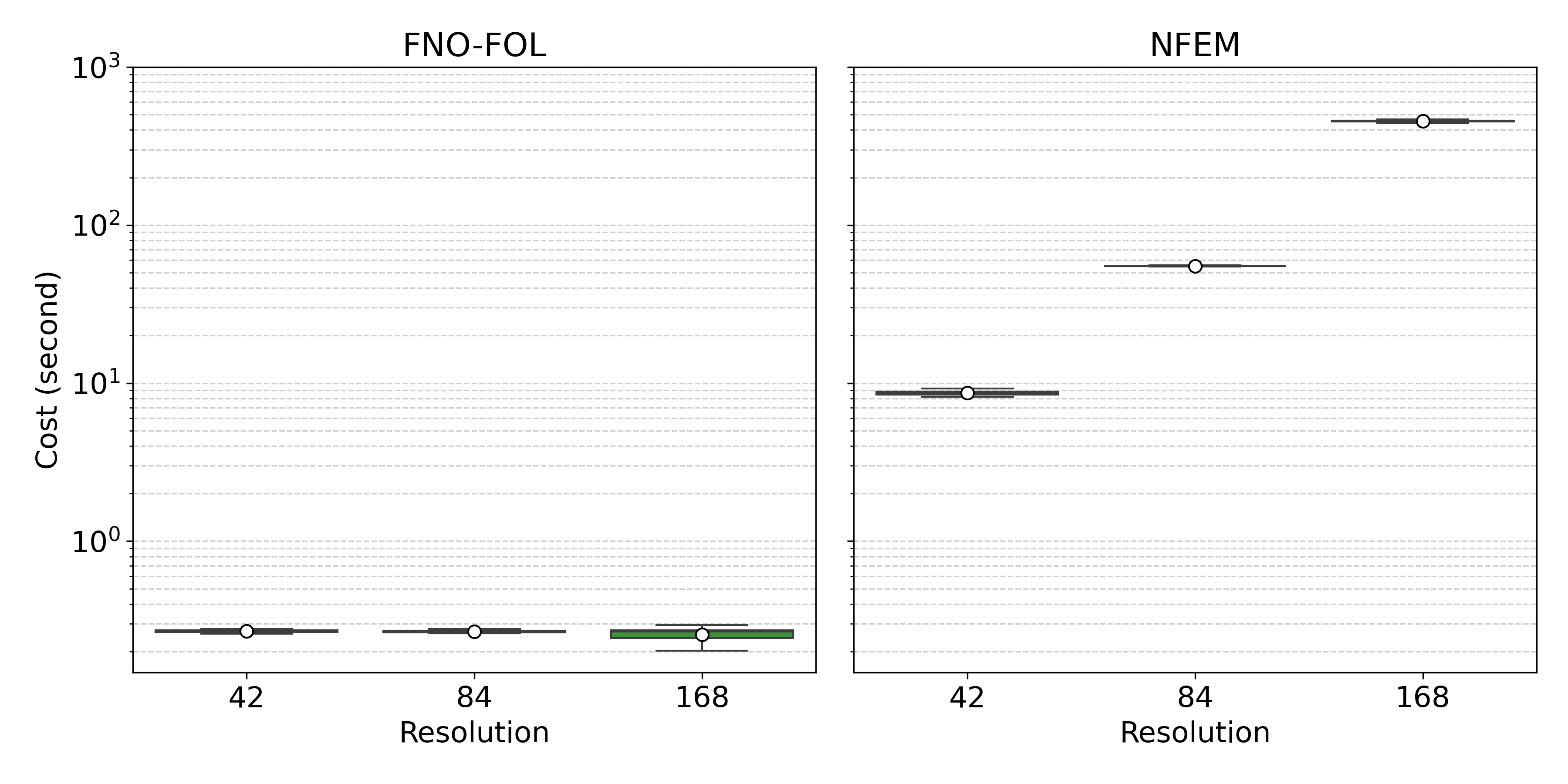}
    \caption{Comparison of average inference cost and calculation cost by NFEM over 50 cases on three different resolutions. }
    \label{fig:timing_2d}
\end{figure}

We evaluate the inference cost of the trained model on three different resolutions, $42 \times 42$, $84 \times 84$, and $168 \times 168$, and compare it with the calculation cost of the conventional nonlinear FEM solver with the Newton-Raphson iterative scheme.
The inference time is measured on a single NVIDIA GeForce RTX 4090 GPU, and the FEM calculation time is also measured on the same GPU using a JAX-based FE solver with a direct linear solver for the Newton-Raphson update.
The results of the measurement over 50 cases are shown in Fig. \ref{fig:timing_2d}. 
It can be observed that the inference cost of the neural operator is on average 32.1 times lower than the calculation cost of FEM for the resolution of $42 \times 42$, 205.0 times lower for the higher-resolution grid of $84 \times 84$, and 1772.3 times lower for the highest-resolution grid of $168 \times 168$, demonstrating the efficiency of the proposed multiphysics FOL-based neural operator for rapid predictions. 
It is also worth noting that the inference cost of the neural operator remains almost constant regardless of the resolution, while the FEM calculation cost increases significantly with higher resolution and higher dimension due to the increased number of degrees of freedom in the nonlinear solver.

\subsection{Three-dimensional nonlinear thermo-mechanics on a Representative Volume Element}
\label{3d_rve}
\begin{figure}[t]
    \centering
    \begin{minipage}{0.45\linewidth}
        \centering
        \includegraphics[width=1.0\linewidth]{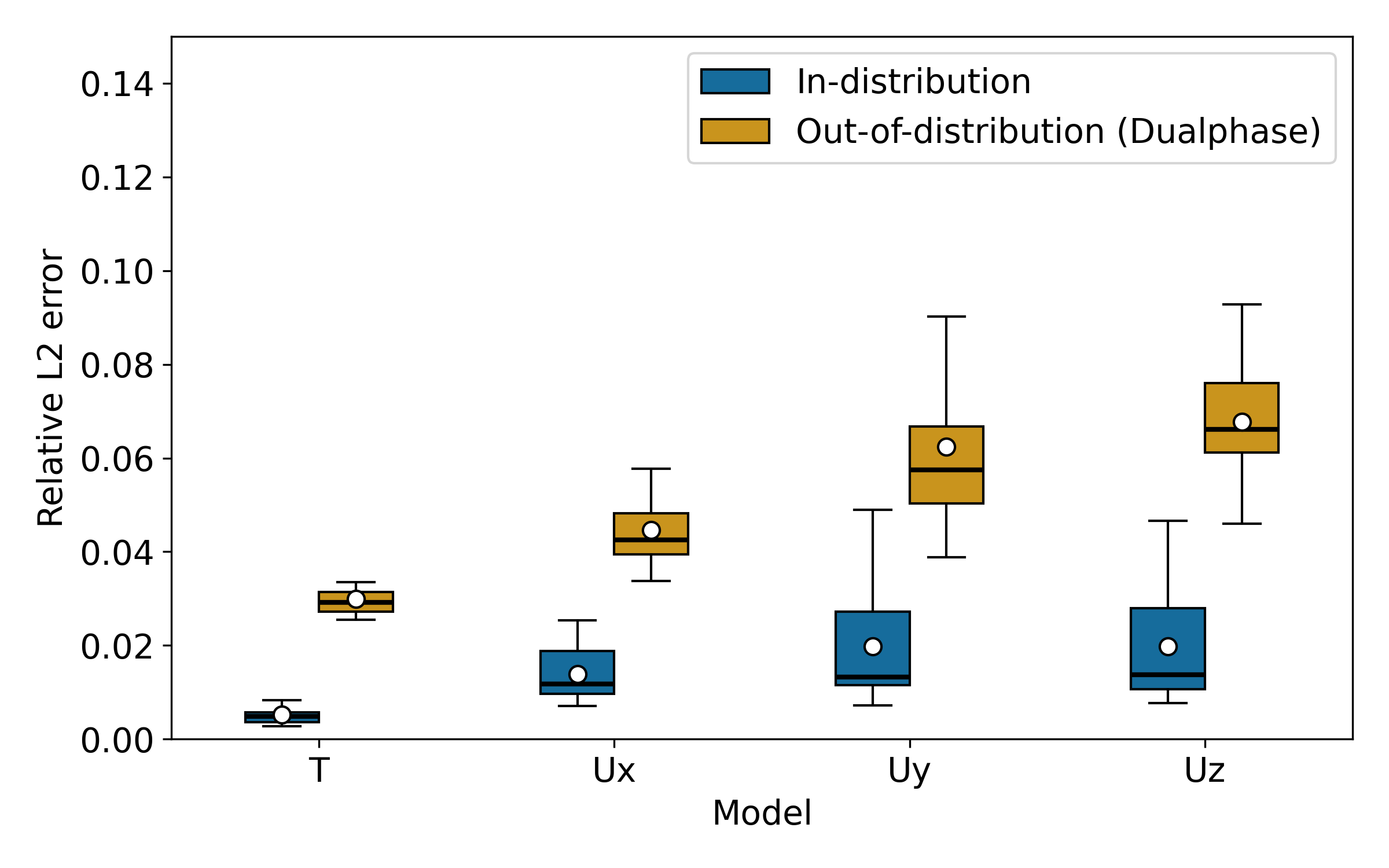}
    \caption{Relative L2 error statistics over 20 in-distribution test samples on the training grid ($22 \times 22 \times 22$) and 20 dualphase test samples on a $44 \times 44 \times 44$ grid.}
    \label{fig:3drve_rel_l2_error}
    \end{minipage}
    \hfill
    \begin{minipage}{0.53\linewidth}
        \centering
        \includegraphics[width=\linewidth]{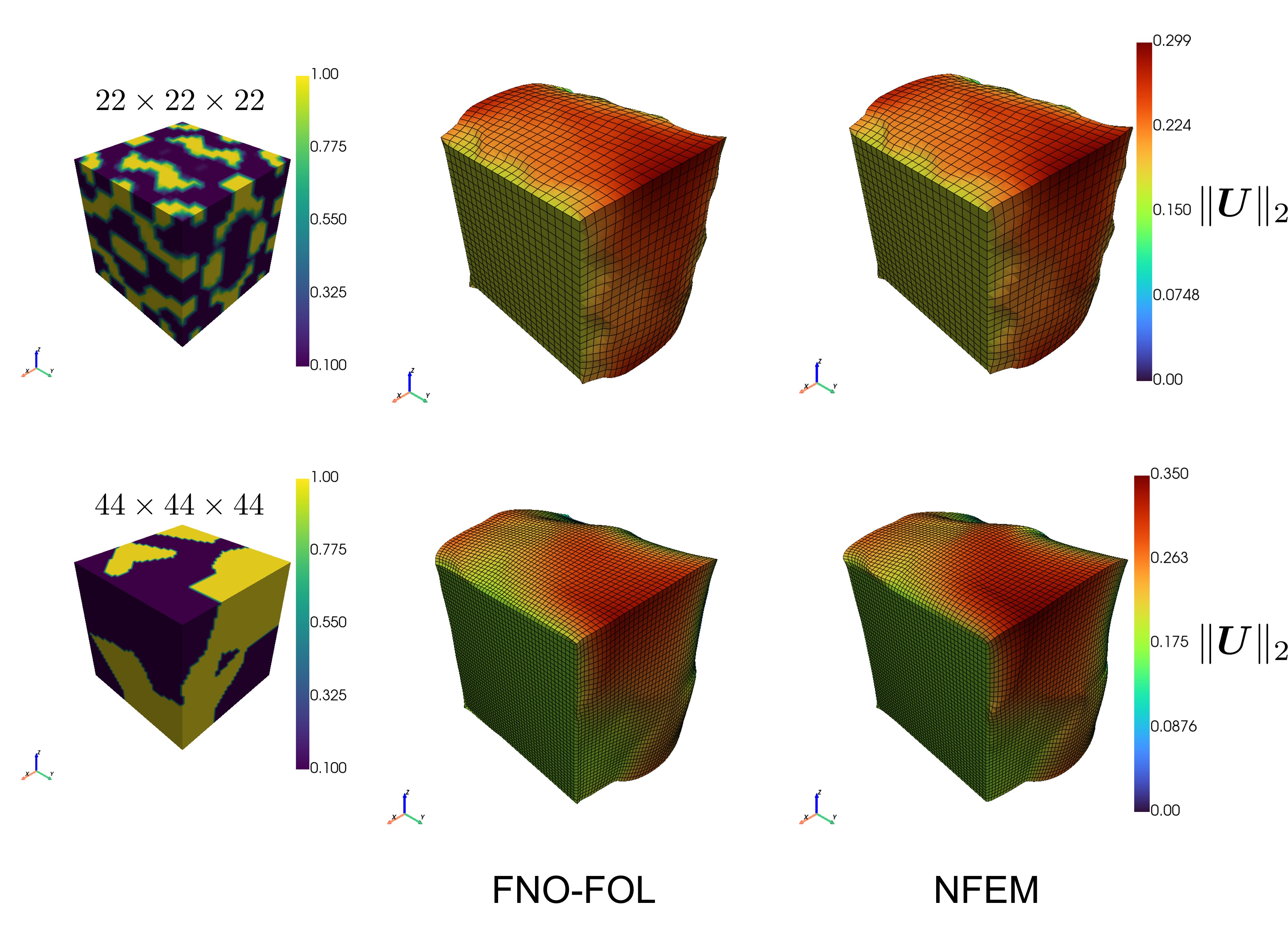}
        \caption{Comparison of the deformed states between prediction and reference solution from NFEM on a representative case of the two test cases.}
    \label{fig:3d_deformation_summary}
    \end{minipage}
\end{figure}

\begin{figure}[H]
    \centering
    \includegraphics[width=0.9\linewidth]{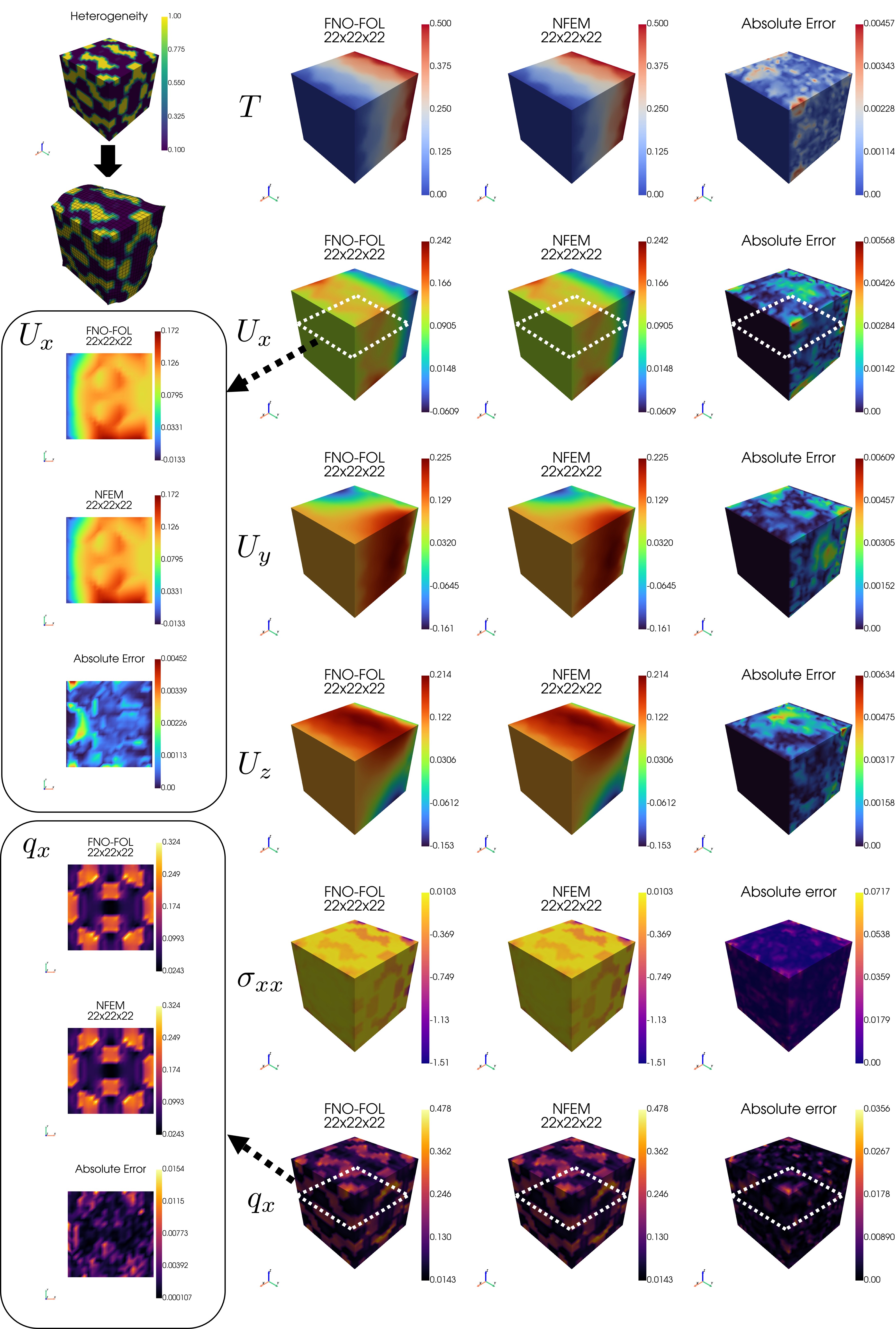}
    \caption{Comparison of the prediction and reference solution from NFEM in the three-dimensional RVE problem for a representative case of the in-distribution test cases.}
    \label{fig:results_3drve_in}
\end{figure}

\begin{figure}[H]
    \centering
    \includegraphics[width=0.90\linewidth]{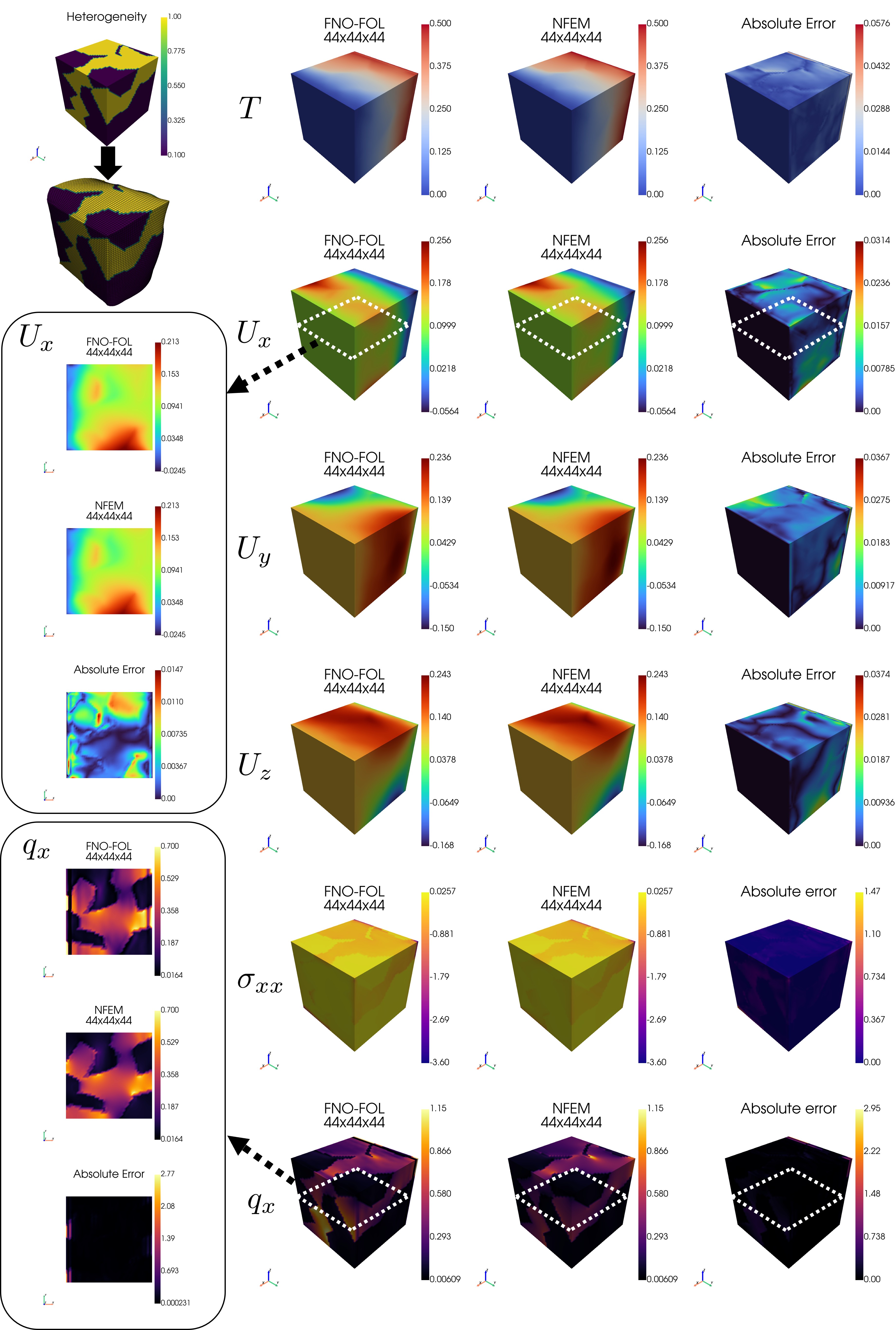}
    \caption{Comparison of the prediction and reference solution from NFEM in the three-dimensional RVE problem for a representative case of the dual-phase structure test cases.}
    \label{fig:results_3drve_dualphase}
\end{figure}

Next, we consider a three-dimensional RVE problem to demonstrate the capability of the proposed multiphysics FOL-based neural operator on three-dimensional problems.
The training hyperparameters are summarized in Table \ref{tab:hyperparameters_fno}, and the geometry and boundary conditions are shown in Fig. \ref{fig:domain_bc} (b) and also in Table \ref{bc_table_3drve}.
The training samples are generated using the Fourier-based function described in \ref{appendix:sample_generation}, similar to the two-dimensional case in Section \ref{2d_square} but expanding the dimensions.
In total, 5000 samples are utilized for training the neural operator model with FNO as the backbone architecture.
The performance of the trained model is evaluated on two different test cases: (1) 20 in-distribution test samples generated using the same Fourier-based function as the training samples, and (2) 20 dual-phase microstructures generated by Voronoi tessellation at a higher resolution of $44 \times 44 \times 44$ to demonstrate the zero-shot super-resolution capability of the trained model.
Here, the dual-phase microstructures are selected as a representative extreme test case since the three extreme test cases in the previous two-dimensional setting do not show significant difference in prediction accuracy between each other.
The relative L2 error statistics over the two test cases are shown in Fig. \ref{fig:3drve_rel_l2_error}. For the in-distribution test samples, the relative L2 error for all physical fields remains below $3\%$ on average, while for the dual-phase test samples, the error remains below $8\%$ on average and $12\%$ even for the largest error, demonstrating the accuracy and generalizability of the proposed multiphysics FOL-based neural operator for three-dimensional problems. The deformed states on a representative sample of the two test cases are compared in Fig. \ref{fig:3d_deformation_summary}, and examples of the prediction results are shown in Figs. \ref{fig:results_3drve_in} and \ref{fig:results_3drve_dualphase}. In the in-distribution test case, FNO-FOL achieves an accurate prediction, which is also confirmed in the cross-section plots on the left of the figure. The dual-phase test case in Fig.~\ref{fig:results_3drve_dualphase} exhibits some aliasing errors around the Dirichlet boundaries, which is analogous to the observation in the two-dimensional case. The error is more pronounced in the heat flux, as seen in the cross-section comparison. However, the overall characteristics of the primary field, stress, and heat-flux responses are accurately captured, suggesting that the predictions are adequate for preliminary inspection of the system behavior as a surrogate model.

The inference cost of the trained neural operator is also evaluated and compared with the calculation cost of the conventional nonlinear FEM solver with the Newton-Raphson iterative scheme.
The inference time is measured on a single NVIDIA GeForce RTX 4090 GPU, and the FEM calculation time is also measured on the same GPU using a JAX-based FE solver with a direct linear solver, similar to the two-dimensional case.
As a result, the inference cost of the neural operator is on average 62.9 times lower than the calculation cost of FEM for the resolution of $22 \times 22 \times 22$ and 638.9 times lower for the higher-resolution grid of $44 \times 44 \times 44$ over the 20 test cases, demonstrating the efficiency of the proposed multiphysics FOL-based neural operator for rapid predictions in three-dimensional problems.

\begin{table}[t]
\centering
\caption{List of temperature and displacement boundary conditions for the three-dimensional nonlinear themo-mechanics on a RVE ; Here, $T$, $U_x$, $U_y$, $U_z$ denote the prescribed temperature or displacements as Dirichlet boundary conditions in the $x$-, $y$-, and $z$-directions respectively, and "free" indicates Dirichlet boundary conditions are not enforced; Instead, natural boundary conditions are applied.}
\begin{tabular}{c|ccccccc}
Location  & $x=0$ & $x=L_x$ & $y=0$ & $y=L_y$ & $z=0$ & $z=L_z$ \\
\hline
$T$  & $0.5 $ & $0.0 $  & free & free & free  & free   \\
$U_x$  & $0.0$ & $0.1 $  & free & free & free  & free   \\
$U_y$ &$0.0$ & $0.1 $  & free & free & free  & free  \\
$U_z$ &$0.0$ & $0.1 $  &free  & free & free & free \\
\hline
\end{tabular}
\label{bc_table_3drve}
\end{table}

\subsection{Nonlinear thermo-mechanics on a three-dimensional casting example}
\label{3d_casting}
\begin{figure}[H]
        \centering
        \includegraphics[width=0.5\linewidth]{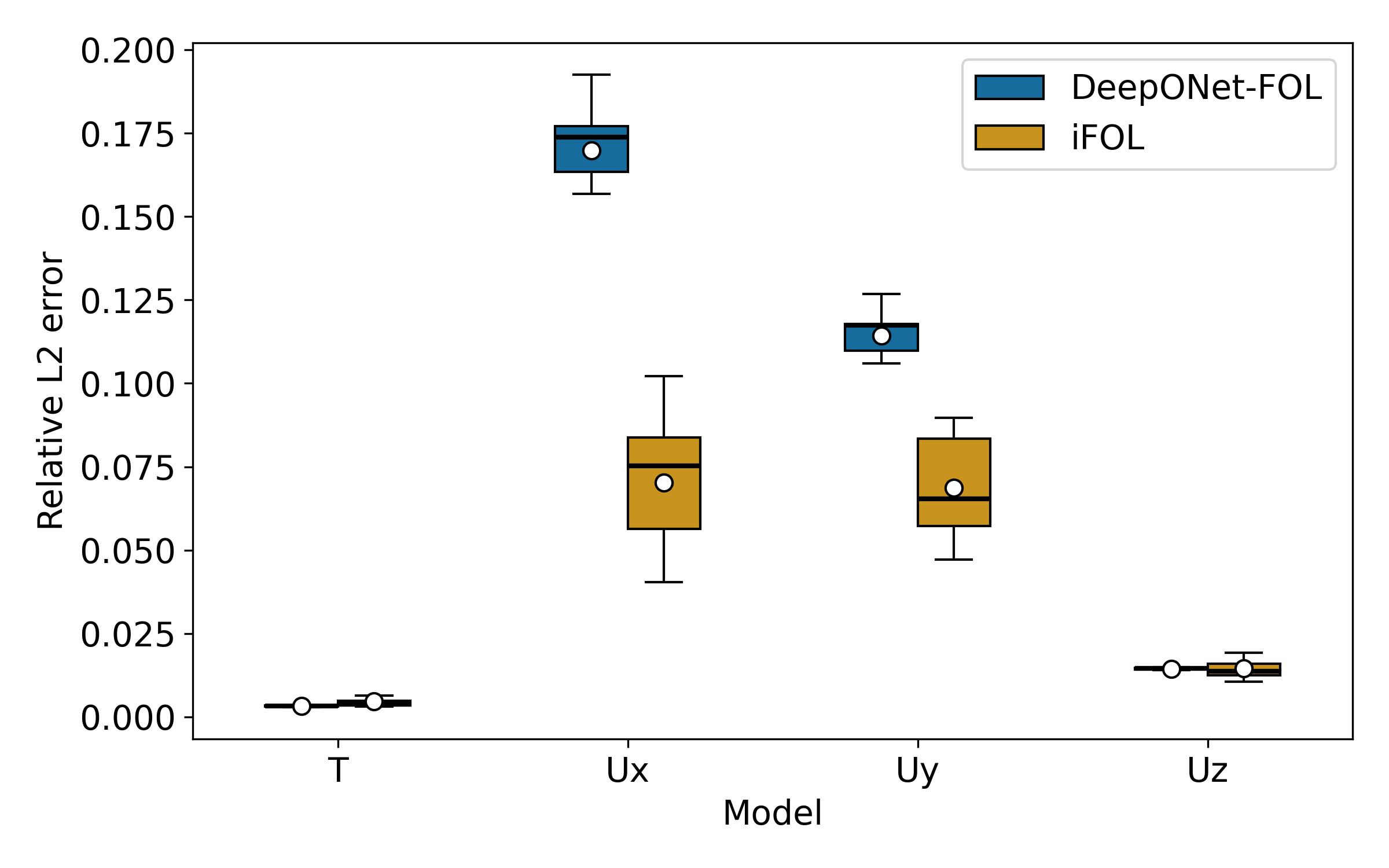}
    \caption{Relative L2 error statistics over 10 uniformly sampled test temperature values on two different operator learning architectures (DeepONet and iFOL) for the irregular casting example geometry.}
    \label{fig:3dcasting_rel_l2}
\end{figure}

\begin{figure}[H]
        \centering
        \includegraphics[width=1.0\linewidth]{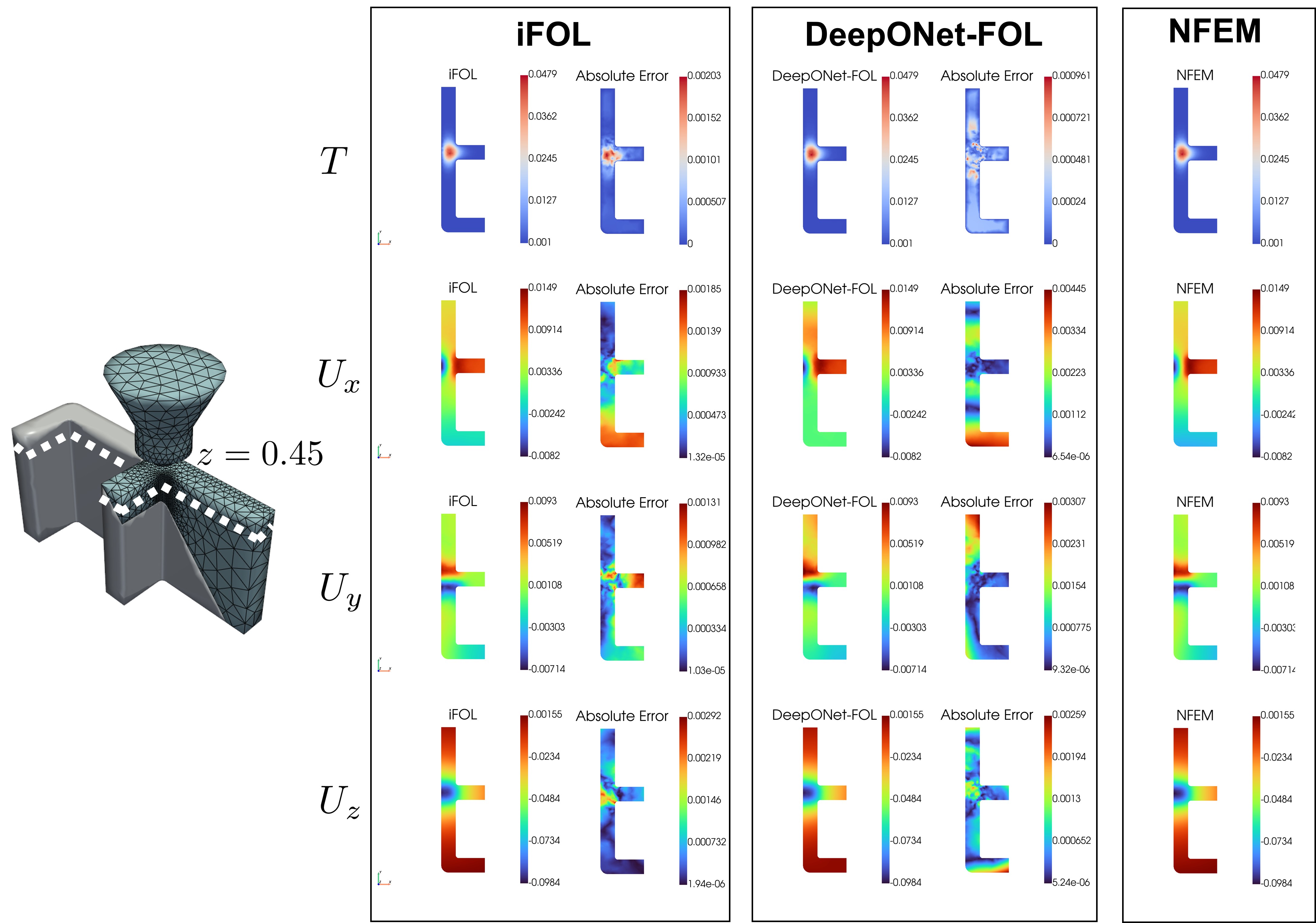}
    \caption{Cross-section comparison of the prediction with iFOL and DeepONet-FOL against the reference NFEM solution at $z=0.45$ in the three-dimensional casting example problem for an unseen temperature case.}
    \label{fig:3dcasting_cross_section}
\end{figure}

\begin{figure}[H]
    \centering
    \includegraphics[width=0.95\linewidth]{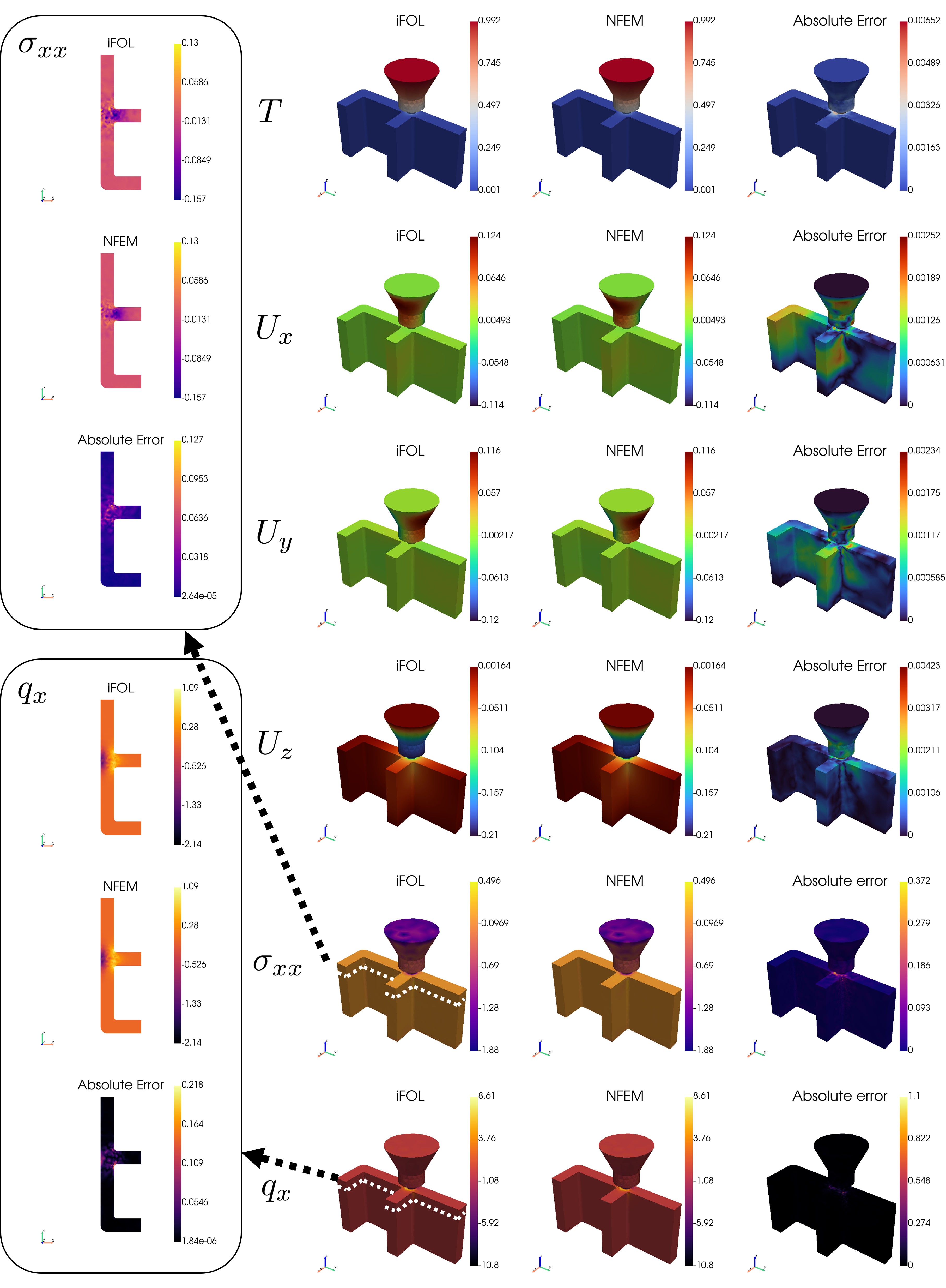}
    \caption{Comparison of the prediction with iFOL and reference solution from NFEM in the three-dimensional casting example problem for an unseen temperature case.}
    \label{fig:results_3dcasting_ifol}
\end{figure}

\begin{figure}[H]
    \centering
    \includegraphics[width=0.95\linewidth]{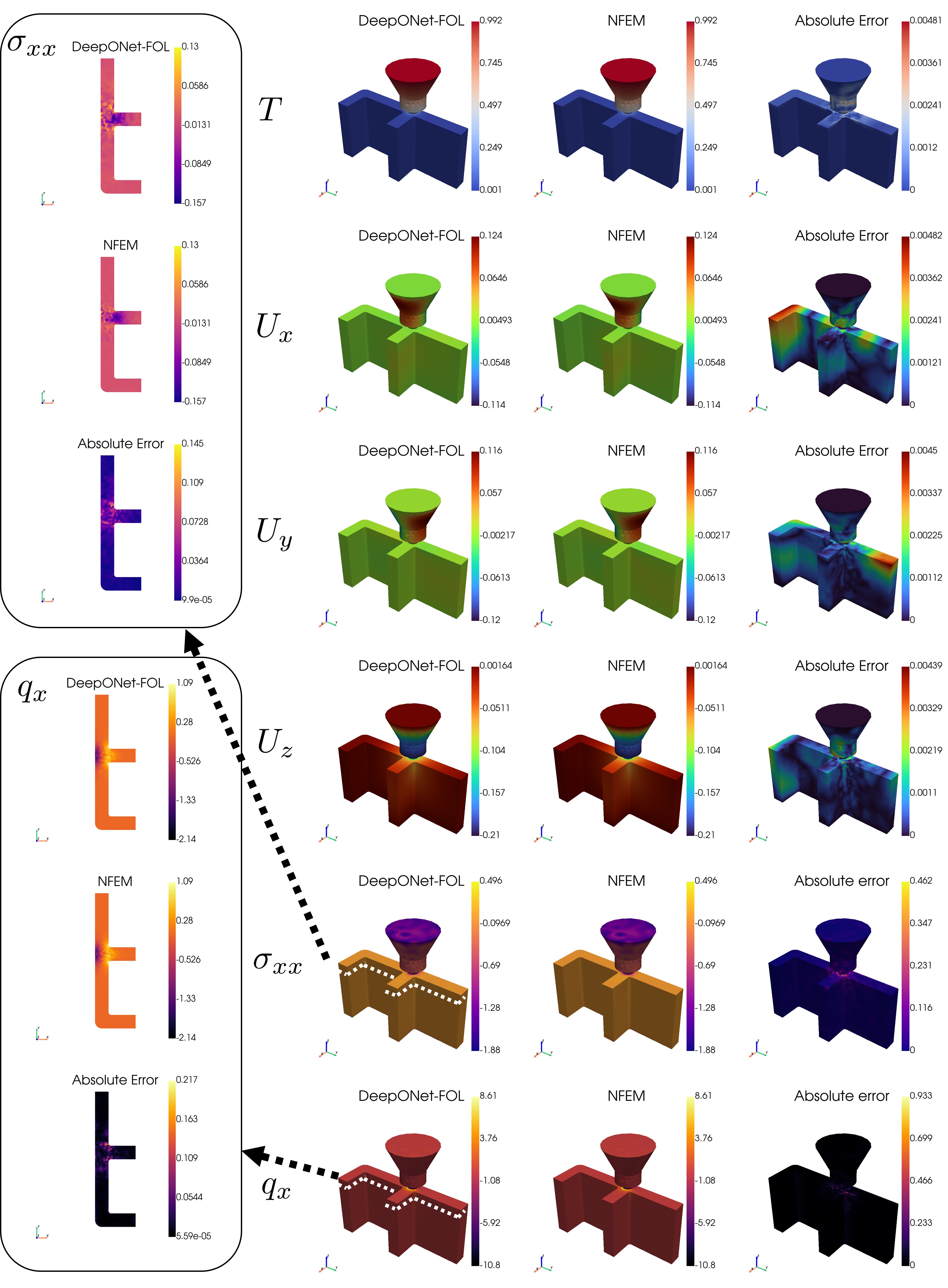}
    \caption{Comparison of the prediction with DeepONet-FOL and reference solution from NFEM in the three-dimensional casting example problem for an unseen temperature case.}
    \label{fig:results_3dcasting_deeponet}
\end{figure}

Lastly, we consider a close-to-reality industrial casting example to demonstrate the capability of the proposed multiphysics FOL-based neural operator on complex and irregular geometries.
The geometry and boundary conditions are shown in Fig. \ref{fig:domain_bc} (c) and Table \ref{bc_table_3dcasting}.
The setting is inspired by a casting process of a mechanical part with an F-shaped geometry, where molten metal is poured from a top inlet and subsequently cools and solidifies.
In this work, we compare two different backbones with FOL, DeepONet and iFOL, to investigate which architecture performs better on complex geometries.  The hyperparameters are summarized in Table \ref{tab:hyperparameters_others}.
The input boundary conditions are parameterized by the temperature value at the top of the feeder in the casting geometry, varying between 0.001 and 1.0 in a normalized scale. For training, 50 values are generated by uniformly sampling the temperature values within the specified range. We limit the number of training samples to 50 to reduce the training time while maintaining a reasonable prediction accuracy upon training.
We train the two models using the same set of training input values and monolithic training scheme. 
The performance of the trained models is evaluated on 10 unseen temperature values uniformly sampled within the same range.
The relative L2 error statistics over the 10 test cases are shown in Fig. \ref{fig:3dcasting_rel_l2}, demonstrating that iFOL outperforms DeepONet-FOL especially in $U_x$ and $U_y$ components.
This could be attributed to the fact that iFOL employs multiple modulator networks acting on each layer of the main neural field network, which allows for more expressive representation of complex mappings compared to the vanilla DeepONet architecture.
The error distributions in Figs. \ref{fig:3dcasting_cross_section}, \ref{fig:results_3dcasting_ifol}, and \ref{fig:results_3dcasting_deeponet} with an unseen temperature of 0.992 given as input also show that iFOL can better capture the spatial distribution of the coupled thermo-mechanical fields compared to DeepONet-FOL, in particular the internal stress response as shown in the top left of Figs.  \ref{fig:results_3dcasting_ifol} and \ref{fig:results_3dcasting_deeponet}.
In this example, it is important to accurately capture the stress distribution since high-stress regions are prone to defects such as cracks during the cooling process in casting. Therefore, the better performance of iFOL in predicting the stress field makes it a more suitable choice for this type of application. 

When comparing the inference cost with NFEM using the bi-conjugate gradient stabilized method for the linear solver due to the efficiency for large-scale problems, iFOL is on average 51.2 times faster than NFEM over the 10 test cases, while DeepONet is on average 81.0 times faster than NFEM. The relatively lower speedup of iFOL compared to DeepONet could be attributed to the additional computational cost introduced by the modulator networks in iFOL. Furthermore, the speedup gained on a fine mesh model is more significant due to the increased computational cost of NFEM with higher degrees of freedom. When comparing the inference cost on a fine mesh model of 55,267 nodes, iFOL is on average 346.3 times faster than NFEM, while DeepONet is on average 595.6 times faster than NFEM. 

\begin{table}[H]
\centering
\caption{List of boundary conditions for the three-dimensional nonlinear themo-mechanics on a casting geometry ; Here, $T$, $U_x$, $U_y$, $U_z$ denote the prescribed temperature or displacements as Dirichlet boundary conditions in the $x$-, $y$-, and $z$-directions respectively, and "free" indicates natural boundary conditions are applied.}
\begin{tabular}{c|cccc}
\hline
Location
& \makecell{Bottom of \\ F-shaped part \\ $z=0.0$}
& \makecell{Surroundings of \\ F-shaped part}
& \makecell{Bottom of \\ inlet part}
& \makecell{Top of \\ inlet part \\ $z=0.85$} \\
\hline
$T$  & $0.001$ & $0.001$ & free &  parametric \\
$U_x$ & $0.0$ & free & $0.0$ & $0.0$ \\
$U_y$ & $0.0$ & free & $0.0$ & $0.0$ \\
$U_z$ & $0.0$ & free & free & $0.0$ \\
\hline
\end{tabular}

\label{bc_table_3dcasting}
\end{table}
\newpage
In summary, these results demonstrate the capability of the proposed FOL-based approach in handling application-oriented complex geometries and boundary conditions under the multiphysics setting, with iFOL showing superior prediction accuracy for this casting example.

\section{Conclusion}
\label{sec:Conclusion}
In this work, we have presented a finite element-based physics-informed neural operator framework for a parametric solution of multiphysics problems.
The proposed framework leverages weighted residuals with the predicted fields serving as the test functions based on the finite element method to construct a physics-informed loss function, which allows for the training of neural operators without the need for labeled data while being capable of seamlessly handling complex geometries and boundary conditions.
We demonstrate the effectiveness of the proposed framework through three numerical examples on a nonlinear thermo-mechanically coupled problem with temperature-dependent material property, including two-dimensional and three-dimesional RVEs, and a three dimensional casting geometry example. 
For the regular domain cases such as RVEs, FNO is employed as the neural operator architecture due to its superior prediction performance through learning on spectral domains, while for the irregular domain case such as the casting geometry, we utilize DeepONet and iFOL as the neural operator architectures due to its ability to handle irregular domains and complex boundary conditions without the need for additional modifications to the architecture. On the two- and three-dimensional RVE cases, the results show that the present framework can accurately predict the solution fields within the error of less than 10\% in relative L2 error even for extreme practical test cases such as gyroid structures, polycrystal structures, and dual-phase polycrystal structures as input. For the three-dimensional casting example case, iFOL is shown to work well in predicting the solution fields with a maximum relative L2 error of 10\% on all the solution fields. This demonstrates its versatility and effectiveness in parametrically predicting solutions to multiphysics problems upon one-time training.
The studies on the aspects of the framework, including the training schemes, network decomposition, and training samples, show that the monolithic training scheme with a single network is sufficient to achieve accurate predictions of the soluton fields, while the training sample quality significantly influences the prediction performance.
It is also worth noting that the proposed framework is highly flexible and can be extended to other multiphysics problems, such as electro-mechanical and chemo-mechanical coupling, by modifying the loss function according to the governing equations of interest. Overall, the current results provide a solid foundation for the development of finite element-guided physics-informed operator learning for multiphysics problems.

Future work could focus on extending the framework to handle transient problems by incorporating time-stepping schemes into the loss function and neural operator architecture. Furthermore, advanced neural operator architectures could be further explored to enable handling multiple input features and addressing varying geometry cases.
Additionally, one could build a multiscale homogenization model by developing hierarchical operator learning architectures that properly incorporate micro-macro coupling.

\section*{CRediT authorship contribution statement}
Yusuke Yamazaki: Writing – review \& editing, Writing – original draft, Conceptualization, Methodology, Software, Visualization, Validation, Investigation, Formal analysis, Data curation.  
Reza Najial Asl: Writing – review \& editing, Conceptualization, Methodology, Software, Validation.
Markus Apel: Writing – review \& editing, Funding aquisition, Resources.
Mayu Muramatsu: Writing – review \& editing, Funding aquisition, Resources. 
Shahed Rezaei: Writing – review \& editing, Writing – original draft, Project administration, Conceptualization, Supervision, Methodology, Software.

\section*{Data availability}
The implementation and the examples presented in this study will be made publicly available upon publication of the work.

\section*{Declaration of competing interest}
The authors declare that they have no known competing financial interests or personal relationships that could have appeared to influence the work reported in this paper

\section*{Declaration of generative AI and AI-assisted technologies in the manuscript preparation process}
During the preparation of this work the authors used ChatGPT and GitHub Copilot in order to assist with drafting and editing the manuscript. After using this tool/service, the authors reviewed and edited the content as needed and take full responsibility for the content of the published article.

\section*{Acknowledgments}
The authors would like to thank the Deutsche Forschungsgemeinschaft
(DFG) for the funding support provided to develop the present work in the Cluster of Excellence Project 'Internet of Production' (project: 390621612). The authors also acknowledge the financial support of SFB 1120 B07 - Mehrskalige thermomechanische Simulation der fest-flüssig Interaktionen bei der Erstarrung (B07) (260070971) (260070971). Yusuke Yamazaki and Mayu Muramatsu acknowledge the financial support from the Japan Society for the Promotion of Science (JSPS) KAKENHI Grant Number JP22H01367. Yusuke Yamazaki acknowledges the financial support from JST BOOST, Japan Grant Number JPMJBS2409.

\appendix
\setcounter{figure}{0}
\setcounter{table}{0}
\section{Hyperparameters of neural operators and simulation parameters}
\label{appendix:hyperparameters}
\begin{figure}[H]
    \centering
    \includegraphics[width=1.0\linewidth]{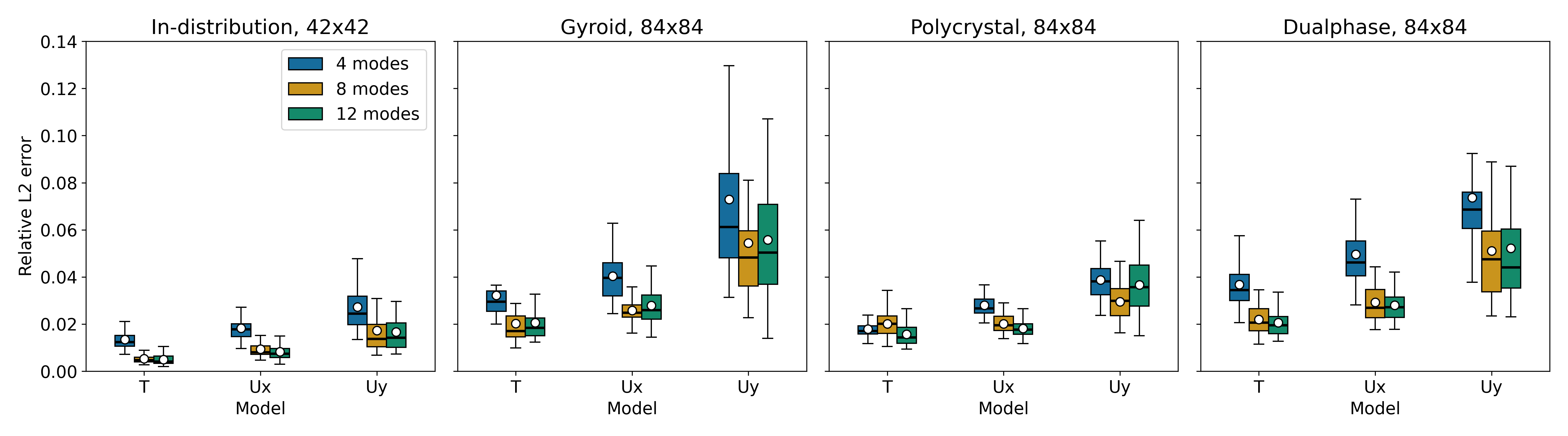}
    \caption{Relative L2 error statistics over 50 samples on four different test cases on different numbers of Fourier modes in FNO.}
    \label{fig:hyper_num_modes}
\end{figure}

\begin{figure}[H]
    \centering
    \includegraphics[width=1.0\linewidth]{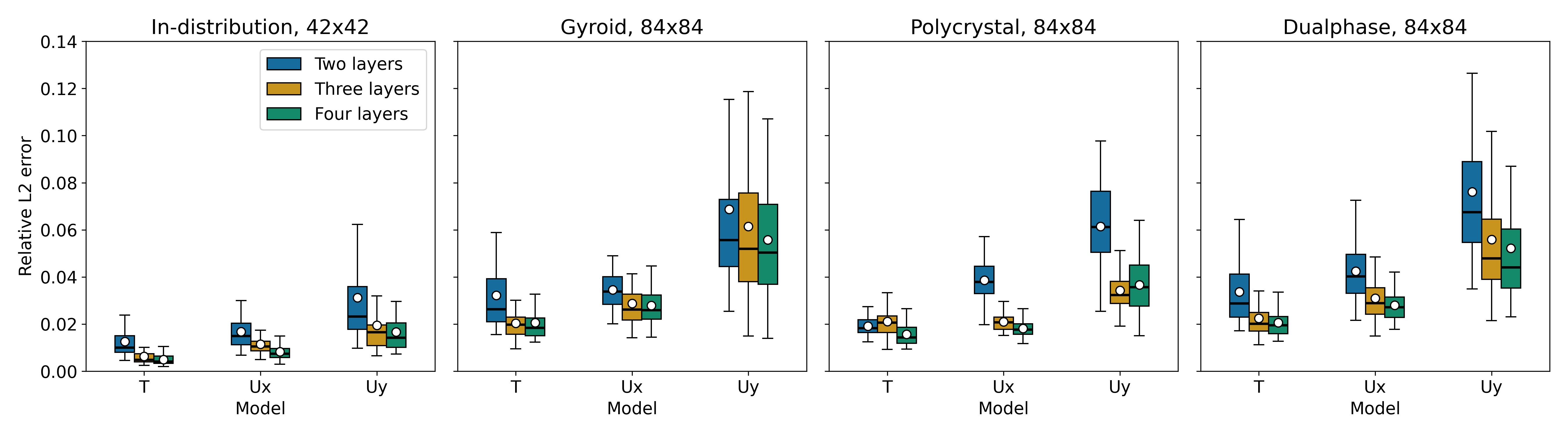}
    \caption{Relative L2 error statistics over 50 samples on four different test cases on different numbers of FNO layers.}
    \label{fig:hyper_layers}
\end{figure}

The list of hyperparameters utilized in this study is summarized in Tables \ref{tab:hyperparameters_fno} and \ref{tab:hyperparameters_others}.
We carefully selected the hyperparameters by carring out hyperparameter studies. Among them, the quantitative comparison on the influence of the number of Fourier modes as well as in FNO for Section \ref{2d_square} are shown in Figures \ref{fig:hyper_num_modes} and \ref{fig:hyper_layers}, respectively. 
The results indicate that the number of Fourier modes and layers significantly affect the prediction accuracy if they are too small. However, after a certain threshold, the accuracy improvement becomes marginal. Based on these observations, we selected 12 modes for the two-dimensional problem and 10 modes for the three-dimensional problem due to the Nyquist criterion, and four FNO layers for both cases.

\begin{table}[H]
\centering
 \caption{Network hyperparameters for the 2D and 3D regular domain problems with Fourier neural operators.
}
 \begin{tabular}{cccc}
    \hline
         Training parameter &  FNO (Sec.~\ref{2d_square}) & FNO (Sec.~\ref{3d_rve}) \\
    \hline
     Number of epochs & 1000 & 1000 \\
     Optimizer & Adam & Adam \\
     Grid in training & $42 \times 42$ & $22 \times 22 \times 22$\\
     Grid in evaluation & $84 \times 84$&  $44 \times 44 \times 44$\\
     Number of modes & (12,12) & (10,10,10) \\
     Number of channels  & 24 & 8 \\  
     Number of Fourier stages & 4 & 4 \\     
     Last projection dimesion & 128 & 128\\
     Output scaling & 0.001 & 0.001 \\ 
     Number of padding for training  & 4 & 4 \\ 
     Activation function & GELU & GELU \\
     Learning rate for main optimization & $1.0 \times 10^{-2} $ to $1.0 \times 10^{-3}$& $1.0 \times 10^{-2} $ to $1.0 \times 10^{-3}$\\
     Number of samples & 5000 & 5000 \\
     Batch size & 100 & 50\\
    \hline
         Total trainable parameters & 2,665,987 & 8,198,436\\\hline\\
    \end{tabular}
\label{tab:hyperparameters_fno}
\end{table}

\begin{table}[H]
\centering
 \caption{Network hyperparameters of DeepONet and iFOL for the three-dimensional casting example problem.
}
 \begin{tabular}{ccc}
    \hline
         Training parameter &   DeepONet (Sec.~\ref{3d_casting}) & iFOL (Sec.~\ref{3d_casting})\\
    \hline
     Number of epochs & 10000& 10000 \\
     Optimizer& Adam& Adam \\
     Number of nodes & 16322& 16322\\
     Branch net/Modulator for each layer & [512,512]& [128,128,128,128] \\  
     Trunk net/Synthesizer & [128,128,128,128]&  [128,128,128,128,128,128]\\  
     Activation function & ReLU & Leaky ReLU \\
     Learning rate for main optimization & $1.0 \times 10^{-3} $ to $1.0 \times 10^{-4}$& $1.0 \times 10^{-3} $ to $1.0 \times 10^{-4}$\\
     Number of samples & 50 & 50 \\
     Batch size & 1 & 1\\
    \hline
         Total trainable parameters & 642,432 & 483,344 \\\hline\\
    \end{tabular}
\label{tab:hyperparameters_others}
\end{table}

\section{Training sample generation}
\label{appendix:sample_generation}
We generate the heterogeneous material samples used in this study by employing a Fourier-based random field generator, as described in \cite{asl2026physics}.
The microstructural patterns are constructed using a phase function $\phi(\boldsymbol{X})$, which varies spatially over the domain. The phase function is generated by superimposing multiple Fourier modes with random amplitudes and frequencies. The general form of the Fourier series used to construct the initial phase function $\phi_f(x,y)$ in a two-dimensional domainis given by:
\begin{equation}
\label{eq:foruier_eq}
\begin{aligned}
    \phi_f(x,y) &= \sum_i^{n_{sum}} [c_i + A_i \sin{(f_{x,i}~x)} \cos{(f_{y,i}~y)} + B_i \cos{(f_{x,i}~x)} \sin{(f_{y,i}~y)} \\ & + C_i\sin{(f_{x,i}~x)} \sin{(f_{y,i}~y)} + D_i \cos{(f_{x,i}~x)} \cos({f_{y,i}~y)}].
\end{aligned}
\end{equation}

Here, $c_i$ denotes a real-valued constant, while $\{A_i, B_i, C_i, D_i\}$ are the amplitudes associated with the corresponding frequency mode. The quantities $\{f_{x,i}, f_{y,i}\}$ represent the frequencies in the $x$- and $y$-directions, respectively.  
To enhance realism and generate more intricate microstructural patterns, we further introduce a phase function $\phi$, obtained through a sigmoidal projection:

\begin{equation}
\label{eq:sigmoid}
\begin{aligned}
    \phi(x,y) = (\phi_{max}-\phi_{min})\cdot\text{Sigmoid}\left(\beta(\phi_f-0.5)\right) + \phi_{min}. 
\end{aligned}
\end{equation}
The above projection ensures that the phase values remain bounded between $\phi_{\min}$ and $\phi_{\max}$. The parameter $\beta$ controls the sharpness of the transition between the two phases.  
For simplicity, the function used in this study retains only the constant term together with the final term involving the product of two cosine functions, i.e., $A_i = B_i = C_i = 0$.
For the two-dimensional studies in Section~\ref{2d_square}, we consider four sets of three frequencies for each direction.
The frequencies in the $x$- and $y$-directions are sampled independently as
\begin{equation}
f_{x,i} \in \{2, 4, 6\}, \{1, 2, 3\}, \{3, 4, 5\}, \{4, 6, 8\},
\quad
f_{y,i} \in \{2, 4, 6\}, \{1, 2, 3\}, \{3, 4, 5\}, \{4, 6, 8\}.
\end{equation}
The Fourier coefficients are sampled from a standard normal distribution, and then the resulting fields are normalized between the specified minimum and maximum values. In this case, we set the minimum as 0.1 and the maximum as 1.0. 
In Section \ref{sec:training_samples_study}, we also investigate the influence of the variety of training samples. To that end, we consider a single set of frequencies as $f_{x,i} \in \{2,4,6\}$ and $f_{y,i} \in \{2,4,6\}$ to generate less-varied training samples.

Including the constant term, this construction yields $M = 3 \times 3 + 1 = 10$ basis terms for each set.
These terms can be combined to construct the phase function $\phi(x,y)$ according to Eq.~\ref{eq:foruier_eq} and Eq.~\ref{eq:sigmoid}. 
In all heterogeneous materials considered in this study, temperature-dependent material properties are defined through the phase function $\phi(\boldsymbol{X})$. For example, in Sections~\ref{2d_square} and \ref{3d_rve}, the thermal conductivity and Young’s modulus are prescribed as the spatially varying material properties $k_0(\boldsymbol{X})= E(\boldsymbol{X}) = \phi(\boldsymbol{X})$. 
As a result, the minimum and maximum stiffness and thermal conductivity values are normalized based on the $\phi(\boldsymbol{X})$. 
For test cases involving polycrystalline materials or other microstructures with varying phase contrasts, the distribution of $\phi(\boldsymbol{X})$ is chosen in an application-dependent manner.
It should be noted, however, that during training we restrict the dataset to samples with a fixed phase contrast ratio, i.e., $\phi_{\max}/\phi_{\min}$.
The extension of this formulation to three dimensions is straightforward; see also \cite{harandi2025spifol}.

\bibliography{references}  
\end{document}